\documentclass{article}
\usepackage{charter,fullpage}
\usepackage{amsmath,amssymb,amsbsy}
\usepackage{multirow}
\usepackage{graphicx}
\usepackage{natbib}
\usepackage{tikz,makecell}
\usetikzlibrary{bayesnet}
\usepackage{hyperref}
\usepackage{textcomp}
\usepackage{subfigure}

\newcommand{\argmin}{\mbox{arg\,min}}

\newcommand{\reals}{\mathbb{R}}
\newcommand{\Prob}{\mathrm{P}}
\newcommand{\E}[1]{\mathbb{E}\left[ #1\right]}

\renewcommand{\gamma}{\mathcal{G}}
\newcommand{\Dir}{\mathrm{Dir}}
\newcommand{\Mult}{\mathrm{Mult}}

\newcommand{\abr}[1]{.}
\newcommand{\mo}[0]{*}

\renewcommand{\b}[1]{\textbf{#1}}
\newcommand{\bs}[1]{\boldsymbol{#1}}
\newcommand{\frob}[1]{\left|\left|{#1}\right|\right|_\mathrm{F}}
\newcommand{\normal}{\mathcal{N}}
\newcommand{\D}{\mathrm{Dir}}
\newcommand{\EF}{\mathrm{EF}}

\usepackage{xcolor}

\allowdisplaybreaks

\begin{document}

\title{Probabilistic Archetypal Analysis}
\author{Sohan Seth and Manuel J. A. Eugster\\
  Helsinki Institute for Information Technology HIIT\\
  Department of Information and Computer Science,
  Aalto University, Finland
}

\maketitle

\begin{abstract}
Archetypal analysis represents a set of observations as convex combinations of
pure patterns, or archetypes. The original geometric formulation of finding
archetypes by approximating the convex hull of the observations assumes them to
be real valued. This, unfortunately, is not compatible with many practical
situations. In this paper we revisit archetypal analysis from the basic
principles, and propose a probabilistic framework that accommodates other
observation types such as integers, binary, and probability vectors. We
corroborate the proposed methodology with convincing real-world applications on
finding archetypal winter tourists based on binary survey data, archetypal
disaster-affected countries based on disaster count data, and document
archetypes based on term-frequency data. We also present an appropriate
visualization tool to summarize archetypal analysis solution better.
\end{abstract}

\section{Introduction}

Archetypal analysis (AA) represents observations as composition of pure
patterns, i.e., \emph{archetypes}, or equivalently convex combinations of
extreme values \citep{cutler_adele_archetypal_1994}.
Although AA bears resemblance with many well established prototypical analysis
tools, such as principal component analysis
\citep[PCA,][]{DBLP:conf/nips/MohamedHG08}, non-negative matrix factorization
\citep[NMF,][]{DBLP:journals/neco/FevotteI11}, probabilistic latent semantic
analysis \citep{hofmann_probabilistic_2013}, and $k$-means
\citep{Steinley@2006}; AA is arguably unique, both conceptually and
computationally. Conceptually, AA imitates the human tendency of representing a
group of objects by its extreme elements \citep{Davis+Love@2010}: this makes AA
an interesting exploratory tool for applied scientists
\citep[e.g.,][]{Eugster@2012,Seiler+Wohlrabe@2013}.  Computationally, AA is
\emph{data-driven}, and requires the
\emph{factors} to be probability vectors: these make AA a
computationally demanding tool, yet brings better interpretability.

The concept of AA was originally formulated by
\citet{cutler_adele_archetypal_1994}. The authors
posed AA as the
problem of learning the convex hull of a point-cloud, and solved it
using alternating non-negative least squares method.  In recent years,
different variations and algorithms based on the original geometrical
formulation have been presented \citep{Bauckhage+Thurau@2009, Eugster+Leisch@2011, morup_archetypal_2012}. However, unfortunately, 
this framework does not tackle many interesting situations. For
example, consider the problem of finding archetypal response to a
binary questionnaire. This is a potentially useful problem in areas of
psychology and marketing research that cannot be addressed in the
standard AA formulation, which relies on the observations to exist in a
vector space for forming a convex hull. Even when the observations
exist in a vector space, standard AA might not be an appropriate tool
for analyzing it. For example, in the context of learning archetypal
text documents with tf-idf as features, standard AA will be inclined
to finding archetypes based on the volume rather than the content of the
document.

\begin{figure*}[t]
  \centering
  \subfigure[]{
    \includegraphics[scale=0.8, clip, trim=0cm 0.3cm 0cm 0.3cm]{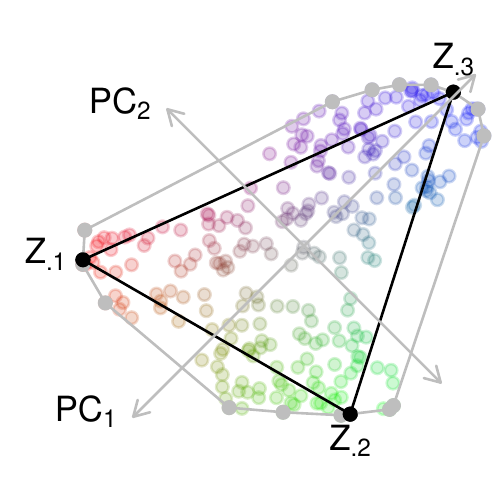}
  }\qquad
  \subfigure[]{
    \includegraphics[scale=0.6, clip, trim=0cm 0cm 0cm 1.5cm]{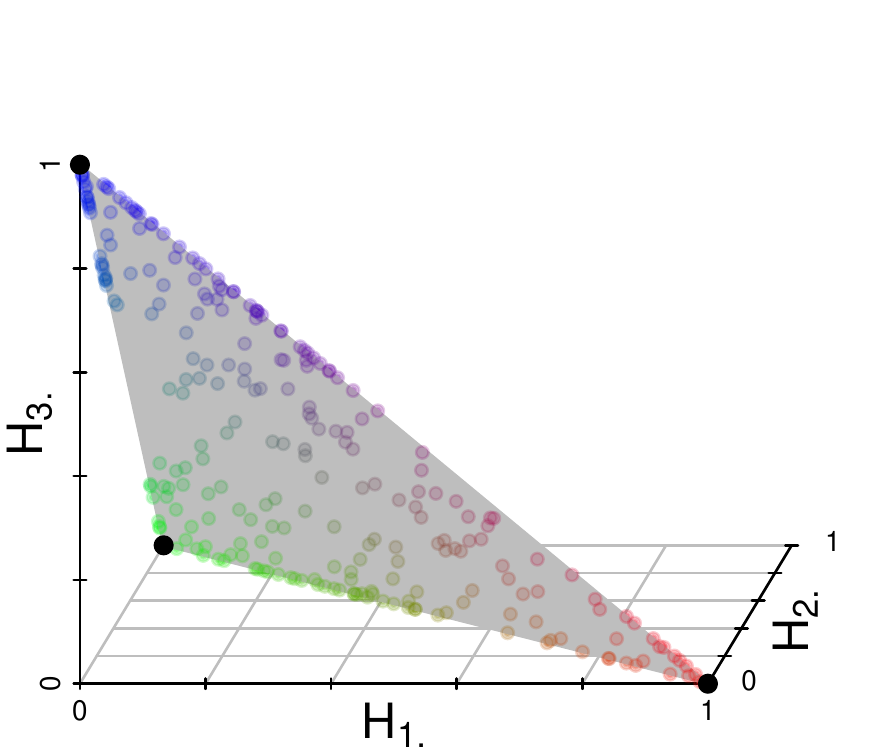}
  }
  \caption{(a)~Illustration of archetypal analysis with three
    archetypes $\b{Z}$, and (b)~the corresponding factors $\b{H}$, projections
of the original observations on the convex hull of the archetypes; (a)~also 
	explicates the difference between PCA and AA.}
  \label{fig:AA}
\end{figure*}

In this paper we revisit archetypal analysis from the basic
principles, and reformulate it to extend its applicability. We admit
that the approximation of the convex hull, as in the standard AA, is
indeed an elegant solution for capturing the essence of `archetypes',
(see Figure~\ref{fig:AA}a and \ref{fig:AA}b for a basic
illustration). Therefore, our objective is to extend the current
framework, not to discard it. We propose a probabilistic foundation of
AA, where the underlying idea is \emph{to form the convex hull in the
parameter space}. The parameter space is often vectorial even if the 
sample space is not (see Figure~\ref{fig:plate} for the plate diagram). 
We solve the resulting optimization problem using majorization-minimization,
and also suggest a visualization tool to help understand the solution
of AA better.

The paper is organized as follows.  In Section~\ref{sec:review}, we
start with an in-depth discussion on what archetypes mean, and how
this concept has evolved over the last decade, and has been utilized
in different contexts.  In Section~\ref{sec:methods}, we provide a
probabilistic perspective of this concept, and suggest
\emph{probabilistic archetypal analysis}. Here we explicitly tackle
the cases of Bernoulli, Poisson and multinomial probability
distributions, and derive the necessary update rules (derivations
available as appendix).  In Section~\ref{sec:foundations}, we discuss
the connection between AA and other prototypical analysis tools---a
connection that has also been partly noted by other researchers
\citep{morup_archetypal_2012}.
In Section~\ref{sec:simulation} we provide simulations to show the
difference betweeen probabilistic and standard archetypal analysis
solutions.
In Section~\ref{sec:vis}, we discuss a visualization method for
archetypal analysis, and present several improvements.  In
Section~\ref{sec:experiments}, we present an application for each of
the above observation models: finding archetypal winter tourists based
on binary survey data; finding archetypal disaster-affected countries
based on disaster count data; and finding document archetypes based on
term-frequency data.  In Section~\ref{sec:discussion} we summarize our
contribution, and suggest future directions.

\begin{table*}[t]
  \centering
  {\normalsize
  \begin{tabular}{llll}
  \hline
  Distribution & Notation & Parameters & pdf/pmf \\
  \hline
  Normal & $\normal(\bs{\mu}, \bs{\Sigma})$ & 
  $\bs{\mu} \in \mathbb{R}^K$, $\bs{\Sigma} \in \mathbb{R}^{K \times K}$
  &
  {\small
  $(2\pi)^{-\frac{K}{2}}|\bs{\Sigma}|^{-\frac{1}{2}} 
    \exp\{-\frac{1}{2}(\bs{x}
    - \bs{\mu})'\bs{\Sigma}^{-1} (\bs{x} - \bs{\mu})\}$} \\

  & & & \\

  Dirichlet & $\D(\bs{\alpha})$ &
  $\bs{\alpha} = (\alpha_1, \ldots, \alpha_K)$, $K > 1$, $\alpha_i > 0$
&
  $\frac{1}{\mathrm{B}(\alpha)} \prod^K_{i = 1} x^{\alpha_i - 1}_i$ 
	where $\mathrm{B}(\alpha) = \frac{\prod^K_{i = 1}
    \Gamma(\alpha_i)}{\Gamma(\sum^K_{i=1} \alpha_i)}$\\
  & & & \\
 
  Poisson & $\mathrm{Pois}(\lambda)$ &
  $\lambda > 0$ &
  $\frac{\lambda^x}{x!} \exp\{-x\}$ \\

  & & & \\

  Bernoulli & $\mathrm{Ber}(p)$ & $0 < p < 1$ & $p^x (1 - p)^{1-x}$ \\ 
 
  & & & \\

  Multinomial & $\Mult(n, \bf{p})$ & $n > 0$, $\bs{p} = (p_1, \ldots, p_K)$,  $\sum^K_{i=1} p_i = 1$& $\frac{n!}{x_1! \cdots
    x_K!}p_1^{x_1} \cdots p_K^{x_K}$\\

  \hline
  \end{tabular}
  }
  \caption{Distributions used in the paper.}
  \label{tab:distr}
\end{table*}

Throughout the paper, we represent matrices by
boldface uppercase letters, vectors by boldface lowercase letter, and
variables by normal lowercase letter. 
$\b{1}$ denotes the row vector of ones, and
$\b{I}$ denotes the identity matrix. Table~\ref{tab:distr} 
provides the definitions of the distributions used
throughout the paper.
Implementations of the presented methods are available at
\url{http://aalab.github.io/}.

\section{Review} \label{sec:review} 

The goal of archetypal analysis is to find archetypes, `pure' patterns. In
Section~\ref{sec:intuition} we provide some intuition on what these pure
patterns imply. In Section~\ref{sec:mathematics}, we discuss the mathematical
formulation of this archetypal analysis as suggested by
\citet{cutler_adele_archetypal_1994}. In Section~\ref{sec:history} we discuss
how this concept has been utilized since its inception:  here, we point out key
references, important developments, and convincing applications.

\subsection{Intuition}
\label{sec:intuition}

Archetypes are `ideal example of a type'. The word `ideal' does
not necessarily have a qualitative meaning of being `good', but this concept
is mostly subjective. For example, one can consider ideal example
to be a prototype, and other objects to be variations of such prototype. This
view is close to the concept of clustering, where the centers of the clusters
are the prototypes. For archetypal analysis, however, ideal example has a
different meaning. Intuitively it implies that the prototype \emph{can not be
surpassed}, its the purest or the most extreme that can be witnessed. 
A simple example of archetypes are the colors red, blue and green (cf.
Figure~\ref{fig:AA}a) in the RGB color space: any other color can be expressed
as combinations of these ideal colors. Another example 
can be comic book superheros who excel in some unique
characteristics, say speed or stamina or intelligence, more than anybody
with these abilities: they are the archetypal
superheros with that particular ability.  The non-archetypal superheroes, on
the other hand, possess ``many'' abilities that are not extreme. 
It is to be noted that a person with all the abilities to their full
realizations, if exists, is an archetype,
Similarly, if one considers normal humans alongside super-humans then
a person with none of these abilities is also an archetype.
  
In both these examples, the archetypes are rather trivial. If one
represents each color in the RGB space then it is obvious that the unit vectors
R, G and B are pure colors or archetypes. 
Similarly, if one represents every (super-)human
in a two dimensional normalized scale of strength and intelligence, then there are four
extreme instances, and hence  
archetypes are: first and second, person with highest score in either of these attributes and none in the other; third and fourth, person
with highest/lowest score in both these attributes.  However, in reality one
may not observe these attributes directly, but some other features. For
example, one can describe a person with many personality traits, 
such as humor, discipline, optimism, etc., but these characteristics
cannot be measured directly. However, one can prepare a questionnaire 
(or observed variables) that explores these (latent) personality traits. 
From this questionnaire, curious users can attempt to identify archetypal
humans, say an archetypal leader or an archetypal jester. 

Finding archetypal patterns in the observed space is a non-trivial problem.  It
is difficult, in particular, since the archetype itself may not be belong to
the set of observed samples but \emph{should be inferred}; yet, it should also
not be a ``mythological'', but rather something that \emph{might be observed}.
\citet{cutler_adele_archetypal_1994}  suggested a simple yet elegant useful
solution that finds the approximate convex hull of the observations, and define
the vertices as archetypes. This allows individual observations to be best
represented by composition (convex combination) of archetypes, while archetypes
can only be expressed by themselves, i.e., they are the `purest form' or `most
extreme' forms. Although, it is certainly not the most desired solution, since,
the inferred archetypes are restricted to be on the boundary of the convex hull
of the observations, whereas true archetype may be outside; inferring such
archetypes outside the observation hull will require strong regularity
assumptions. The solution suggested by \citet{cutler_adele_archetypal_1994}
finds a trade off between computational simplicity, and the intuitive nature of
archetype. 

\renewcommand{\frob}{\mathrm{F}}
\subsection{Formulation}
\label{sec:mathematics}
\citeauthor{cutler_adele_archetypal_1994} posed AA as the problem
of learning the convex hull of a point-cloud. They assumed
the archetypes to be convex combinations of observations, and the
observations to be convex combinations of the archetypes. 
Let $\b{X}$ be a
(real-valued) data matrix with each column as an observation.  Then, 
this is equivalent to solving the following optimization
problem:
\begin{align}
\label{eq:saa} 
\min_{\b{W},\b{H}} ||\b{X} - \b{X}\b{W}\b{H}||^2_\mathrm{F} 
\end{align} 
with the constraint that both $\b{W}$ and $\b{H}$ are column stochastic
matrices.  $\mathrm{F}$ denotes Frobenious norm. Given $N$ observations, and
$K$ archetypes, $\b{W}$ is $N\times K$ dimensional, and $\b{H}$ is $K \times N$
dimensional matrices. Here, $\b{Z}=\b{X}\b{W}$ are the inferred archetypes that
exist on the convex hull of the observations due to the stochasticity of
$\b{W}$ and for each $n$-th sample $\b{x}_n$,
$\b{Z}\b{h}_n$ is its projection on the convex hull of the archetypes.

\citeauthor{cutler_adele_archetypal_1994} 
solved this problem using an alternating non-negative least squares
method as follows: 
\[ 
\b{H}^{t+1} = \argmin_{\b{H} \geq \b{0}} ||\b{X} -
\b{Z}^{t}\b{H}||_\frob^2 + \lambda||\b{1}\b{H}-\b{1}||^2
\]
and
\[
\b{W}^{t+1} = \argmin_{\b{W} \geq \b{0}} ||\b{Z}^{t} -
\b{X}\b{W}||_\frob^2 + \lambda||\b{1}\b{W}-\b{1}||^2
\]
where after each alternating step, the archetypes ($\b{Z}$) are
updated by solving $\b{X} = \b{Z}^{t+1}\b{H}^{t}$, and $\b{Z}^{t+1} =
\b{X}\b{W}^t$, respectively. 
The algorithm alternates between finding the best composition
of observations given a set of archetypes, and then finding the best
set of archetypes given a composition of observations. 
Notice that the stochasticity constraint
was cleverly enforced by a suitably strong regularization parameter
$\lambda$.  The authors
also proved that $k > 1$ archetypes are located on the boundary of the
convex hull of the point-cloud, and $k = 1$ archetype is the mean of
the point cloud. 

\subsection{Development}
\label{sec:history}

The first publication, to the best of our knowledge, which deals with the
idea of ``ideal types'' and observations related to them, is
\citet{woodbury_clinical_1974}. There, the authors discuss how to
derive estimates of grades of membership of categorical observations,
given an a-priori defined set of ideal (or pure) types, in the context
of clinical judgment. Twenty years later---in
1994---\citet{cutler_adele_archetypal_1994} formulated archetypal
analysis~(AA) as the problem of estimating both the membership and the
ideal types given a set of real-valued observations. They motivated
this new kind of analysis with, among other examples, the estimation
of archtyepal head dimensions of Swiss Army
soldiers. 

One of the original authors continued her work on AA in the fields
of physics and applied it on spatio-temporal data
\citep{stone_archetypal_1996}. In this line of research,
\citet{cutler_moving_1997} developed \emph{moving archetypes}, by extending
the original AA framework with an additional
optimization step, which estimates the optimal shift of observations in
the spatial domain over time. They applied this method to data gathered
from a chemical pulse experiment.
Other researches took up the idea of AA and applied it in different
fields; the following is a comprehensive list of problems where other 
researchers have applied AA: analysis of galaxy
spectra~\citep{chan_archetypal_2003}, ethical issues and market
segmentation~\citep{li_archetypal_2003}, thermogram
sequences~\citep{marinetti_archetypes_2007}, gene expression
data~\citep{thogersen_archetypal_2013}; performance analysis in
marketing~\citep{porzio_use_2008},
sports~\citep{Eugster@2012}, and
science~\citep{Seiler+Wohlrabe@2013}; face
recognition~\citep{xiong_face_2013}; and in game AI development
\citep{sifa_archetypical_2013}.

In recent years, animated by the rise of the non-negative matrix
factorization research, various authors have proposed extensions and
variations to the original algorithm. Following are a few notable publications.
\citet{thurau_convex_2009} introduce the convex-hull non-negative
matrix factorization (NMF). Motivated by the convex NMF, the authors
make the same assumption as made in AA that observations are 
convex combinations of specific observations. However,
they derive an algorithm which estimates the archetypes not from
the entire set of observations but from potential candidates found from
2-dimensional projections on eigenvectors:
this leads to a solution also applicable for large data sets. The
authors demonstrate their method on a data set consisting of
150~million votes on World of Warcraft$^\text{\textregistered}$
guilds. In \citet{thurau_yes_2010}, the authors 
present an even faster approach by deriving a highly efficient volume
maximization algorithm.
\citet{Eugster+Leisch@2011} tackle the problem of robustness, and
that a single outlier can break down the archetype solution. They
adapt the original algorithm to be a robust M-estimator and present an
iteratively reweighted least squares fitting algorithm. They evaluate
there algorithm using the Ozone data from the original AA paper with
contaminated observations.
\citet{morup_archetypal_2012} also tackle the problem of deriving an
algorithm for large scale AA. They propose a solution based on a
simple projected gradient method, in combination with an efficient
initialization method for finding candidates of archetypes. The authors
demonstrate their method, among other examples, with an analysis of
the NIPS bag of words corpus and the Movielens movie rating data set.

\section{Probabilistic archetypal analysis}
\label{sec:methods}


We observe that 
the original AA formulation
implicitly exploits a \emph{simplex latent variable  model}, 
and normal observation
model, i.e., 
\[\b{h}_n\sim\Dir(\b{1}),\,\b{x}_n \sim \normal(\b{Z}\b{h}_n,\epsilon_1\b{I}).\]
But, it goes a step further,
and generates the loading matrix $\b{Z}$ from a simplex latent model
itself with \emph{known} loadings $\bs{\Theta} \in \reals^{M\times N}$, i.e.,
\[\b{w}_k\sim\Dir(\b{1}),\,\b{z}_k \sim \normal(\bs{\Theta}\b{w}_k,\epsilon_2\b{I}).\] 
Thus, the log-likelihood can be written as, 
\begin{align*}
\mathbb{LL}(\b{X}|\b{W},\b{H},\b{Z},\bs{\Theta}) = 
-\frac{\epsilon_1}{2}||\b{X} - \b{Z}\b{H}||_\frob^2 - \frac{\epsilon_2}{2}||\b{Z}
-\bs{\Theta} \b{W}||_\frob^2+ C(\epsilon_1,\epsilon_2). 
\end{align*}
The archetypes $\b{Z}$, and corresponding factors $\b{H}$ can then be
found by maximizing this log-likelihood (or minimizing the negative
log-likelihood) under the constraint that both $\b{W}$ and $\b{H}$ are
stochastic: this can be achieved by alternating optimization as
Cutler and Breiman did (but with different update rules for $\b{Z}$, and
$\epsilon_{\cdot}$). 

The equivalence of this approach to the standard formulation
requires that $\bs{\Theta} = \b{X}$.  Although unusual in a probabilistic
framework, this contributes to the data-driven nature of AA. In the
probabilistic framework, $\bs{\Theta}$ can be viewed as a set of known
bases that is defined by the observations, and the purpose of archetypal
analysis is to find a \emph{sparse} set of bases that can explain the observations.
These inferred bases are the archetypes, and the stochasticity constraints
on $\b{W}$ and $\b{H}$ ensure that they are the extreme values as one desires. 
It should be noted that 
$\bs{\Theta}_n$ does not need to correspond to $\b{X}_n$: more generally,
$\bs{\Theta} = \b{X}\b{P}$ where $\b{P}$ is a permutation matrix.
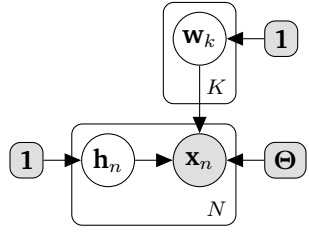
\begin{figure}[t]
 \centering
 \begin{tikzpicture}[font=\normalfont]
    \node[rectangle, draw=black,fill=gray!25, rounded corners] (c) {$\b{1}$};
    \node[latent, right=0.5cm of c] (x) {$\b{h}_n$};
    \node[obs, right=0.5cm of x] (y) {$\b{x}_n$};
    \node[rectangle, right=0.5cm of y, draw=black,fill=gray!25, rounded corners] (Y) {$\bs{\Theta}$};
    \node[latent, above=0.9cm of y] (z) {$\b{w}_k$};
    \node[rectangle, draw=black,fill=gray!25, rounded corners, right=0.5cm of z] (d) {$\b{1}$};

    \edge {c}{x};
    \edge {x}{y};
    \edge {Y}{y};
    \edge {z}{y};
    \edge {d}{z};

    \plate {xy}{(x)(y)}{$N$};
    \plate {z}{(z)}{$K$};
  \end{tikzpicture}
  \caption{Plate diagram of probabilistic archetypal analysis.}
  \label{fig:plate}
\end{figure}

\subsection{Exponential family} 
We describe AA in a probabilistic set-up as follows (see Figure~\ref{fig:plate}),
\[
\b{w}_k \sim \D(\b{1}), \,\b{h}_n \sim \D(\b{1}),\, \b{x}_n \sim
\EF(\b{x}_n;\bs{\Theta}\b{W}\b{h}_n) 
\]
where
\[
\EF(\b{z};\bs{\theta}) =
h(\b{z})g(\bs{\theta})\exp(\eta(\bs{\theta})^\top s(\b{z}))
\]
with standard meaning for the functions $g,h,\eta$ and $s$.
Notice that, \emph{we employ the normal
parameter $\bs{\theta}$ rather than the natural parameter
$\bs{\eta}(\bs{\theta})$}, since the former is more interpretable. In
fact, the convex combination of $\bs{\theta}$ is more interpretable
than the convex combination of $\bs{\eta}(\bs{\theta})$, as a linear
combination on $\bs{\eta}(\bs{\theta})$ would lead to nonlinear
combination of $\bs{\theta}$. 
To adhere to the original formulation, we suggest $\bs{\Theta}_{\cdot n}$ 
to be the \emph{maximum likelihood point estimate} from
observation $\b{X}_{\cdot n}$. Again, the columns of $\bs{\Theta}$ and $\b{X}$
do not necessarily have to be corresponded.  
Then, we find archetypes $\b{Z}=\bs{\Theta}\b{W}$
by solving
\begin{align}\label{eq:paa}
\argmin_{\b{W},\b{H} \geq 0} -\mathbb{LL}
(\b{X}|\b{W},\b{H},\bs{\Theta})\mbox{ such that } \b{1}\b{W}=\b{1},
\b{1}\b{H} = \b{1}.
\end{align}
We call this approach \emph{probabilistic archetypal analysis} (PAA).

The meaning of \emph{archetype} in PAA is different than in the
standard AA since the former lies in the parameter space,
whereas the latter in the observation space. To differentiate these
two aspects, we call the archetypes $\b{Z}=\bs{\Theta}\b{W}$ found by PAA
(solving \eqref{eq:paa}), \emph{archetypal
profiles}: our motivation is that $\bs{\Theta}_{\cdot n}$
can be seen as the parametric \emph{profile} that best describes the
single observation $\b{x}_n$, and thus, $\b{Z}$ are the archetypal profiles
that are inferred from them. 
We generally refer to the set of indices
that contribute to the $k$-th archetypal profile, i.e., 
$\{i : \b{W}_{ik} > \delta\}$, where $\delta$ is a small value,
as \emph{generating observations} of that archetype.
Notice that, when the observation model is multivariate
normal with identity covariance, then this formulation is the same as
solving \eqref{eq:saa}. We explore some other examples of $\EF$:
multinomial, product of univariate Poisson distributions, and product
of Bernoulli distributions.

\subsection{Poisson observations}

If the observations are integer valued then they are usually assumed
to originate from a Poisson distribution. Then we need to solve the following problem,
\[
\argmin_{\b{W},\b{H} \geq \b{0}} \sum_{mn} \left[
  -\b{X}_{mn}\log(\bs{\Lambda} \b{W} \b{H})_{mn} + (\bs{\Lambda} \b{W}
  \b{H})_{mn}\right]
\]
such that $\sum_{j}\b{H}_{jn} = 1 \mbox{ and } \sum_i \b{W}_{ik} = 1$.
Here $\Lambda_{mn}$ is the maximum likelihood estimate of the 
Poisson rate parameter from observation $\b{X}_{mn}$.

To solve this problem efficiently, we employ a similar technique used
by Cutler and Breiman by relaxing the equality constraint with a
suitably strong regularization parameter. However, we employ a
multiplicative update rule afterwards instead of an exact method like
the nonnegative least squares. The resulting update rules are 
(see Appendix~\ref{sec:pois} for derivation)
\begin{align*}
\b{H}^{t+1} &= \b{H}^t \odot
\frac{\nabla_{\b{H}^t}^-}{\nabla_{\b{H}^t}^+}, \,\nabla_{\b{H}_{nj}}^+
= \sum_{im} \bs{\Lambda}_{im} \b{W}_{mn} + \lambda, \nabla_{\b{H}_{nj}}^- \\
&= \sum_{i} \frac{\b{X}_{ij} \sum_{m} \bs{\Lambda}_{im}
  \b{W}_{mn}}{\sum_{mn } \bs{\Lambda}_{im} \b{W}_{mn} \b{H}_{nj}}  +
\frac{\lambda}{\sum_n \b{H}_{nj}}
\end{align*}
and
\begin{align*}
\b{W}^{t+1} &= \b{W}^t \odot
\frac{\nabla_{\b{W}^t}^-}{\nabla_{\b{W}^t}^+}, \,\nabla_{\b{W}_{mn}}^+
= \sum_{ij} \bs{\Lambda}_{im} \b{H}_{nj} + \lambda, \nabla_{\b{W}_{mn}}^- \\
&= \sum_{ij} \frac{\b{X}_{ij} \bs{\Lambda}_{im} 
\b{H}_{nj}}{\sum_{mn } \bs{\Lambda}_{im} \b{W}_{mn} \b{H}_{nj}} +
\frac{\lambda}{\sum_m\b{W}_{mn}}\mbox{.}
\end{align*}
Here $\odot$ denotes Hadamard product. We choose $\lambda$ to be 20
times the variance of the samples.

\subsection{Multinomial observations}

In many practical problems such as document analysis, the observations
can be thought of as originating from a multinomial model. In such
cases, PAA expresses the underlying multinomial probability as
$\b{P}\b{W}\b{H}$ where $\b{P}$ is the maximum likelihood estimate
achieved from word frequency matrix $\b{X}$. This decomposition is
very similar to PLSA: PLSA estimates a topic by document matrix
$\b{H}$ and a word by topic matrix $\b{Z}$, while AA estimates a
document by topic matrix ($\b{W}$) and a topic by document matrix
($\b{H}$) from which the topics can be estimated as archetypes
$\b{Z}=\b{P}\b{W}$.  Therefore, the archetypal profiles are
effectively topics, but topics might not always be archetypes. For
instance, given three documents \{A,B\}, \{B,C\}, \{C,A\}; the three
topics could be \{A\}, \{B\}, and \{C\}, whereas the archetypes can
only be the documents themselves. Thus, it can be argued that
archetypes are topics with better interpretability.

To find archetypes for this observation model one needs to solve the
following problem, 
\begin{align*}
\argmin_{\b{W},\b{H} \geq \b{0}} -\sum_{mn}\b{X}_{mn}
\log{\sum_{ij}\b{P}_{mi}\b{W}_{ij}\b{H}_{jn}},
\end{align*}
such that $\sum_{j}\b{H}_{jn} = 1$ and $\sum_i \b{W}_{ik} = 1$.

This can be efficiently solved using expectation-maximization (or
majorization-minimization) framework with the following update rules
(see Appendix~\ref{sec:multi} for derivation),
\begin{align*}
\b{H}_{ij}^{t+1} =
\sum_{kl}\frac{\b{X}_{il}\b{H}_{ij}\b{W}_{jk}\b{P}_{kl}}{(\b{H}\b{W}\b{P})_{il}},\;
\b{H}_{ij}^{t+1} = \frac{\b{H}_{ij}^{t+1}}{\sum_{j} \b{H}_{ij}^{t+1}}
\end{align*}
and
\begin{align*}
\b{W}_{jk}^{t+1} =
\sum_{il}\frac{\b{X}_{il}\b{H}_{ij}\b{W}_{jk}\b{P}_{kl}}{(\b{H}\b{W}\b{P})_{il}},\;
\b{W}_{jk}^{t+1} = \frac{\b{W}_{jk}^{t+1}}{\sum_{k} \b{W}_{jk}^{t+1}}\mbox{.}
\end{align*}

\subsection{Bernoulli observations}

There are real world applications that deal with binary observations
rather than real valued or integers, e.g., binary questionnaire in
marketing research.  Such observations can be expressed in terms of
the Bernoulli distribution. To find the archetypal representation of
binary pattern we need to solve the following problem,
\[ 
\argmin_{\b{W},\b{H} \geq \b{0}} \sum_{mn} \left[ - \b{X}_{mn}
  \log(\b{P}\b{W}\b{H})_{mn} - \b{Y}_{mn}
  \log(\b{Q}\b{W}\b{H})_{mn}\right],
\]
such that $\sum_{j}\b{H}_{jn} = 1 \mbox{ and } \sum_i \b{W}_{ik} = 1$,
where $\b{X}$ is the binary data matrix (with $1$ denoting
success/true and $0$ denoting failure/false), $\b{P}_{mn}$ is the
probability of success estimated from $\b{X}_{mn}$ (effectively either
$0$ or $1$),  $\b{Y}_{mn} = 1 - \b{X}_{mn}$, and $\b{Q}_{mn} = 1 -
\b{P}_{mn}$. 

This is a more involved form than the previous ones: one cannot use
relaxation technique as in the Poisson case, since relaxation over the
stochasticity constraint might render the resulting probabilities
$\b{P}\b{W}\b{H}$ greater than $1$, thus making the cost function
incomputable. 
 Therefore, we
take a different approach toward solving this problem by
reparameterizing the stochastic vector (say $\b{s}$) by an
unnormalized non-negative vector (say $\b{t}$), such that $\b{s} =
\b{t}/\sum\b{t}_i$. We show that the structure of the cost allows us
to derive efficient update rules over the unnormalized vectors using
majorization-minimization. Given $\b{g}_n$ and $\b{v}_k$ to be the
reparameterization of $\b{h}_n$ and $\b{w}_k$ respectively, we get the
following update equations,
(see Appendix~\ref{sec:bern} for derivation),
\[
\b{G}^{t+1} = \b{G}^t \odot \frac{\nabla_{\b{G}}^n}{\nabla_{\b{G}}^d}\mbox{,}
\]
with
\begin{align*}
{\nabla_{\b{G}_{nj}}^d} = &\sum_i \b{X}_{ij} + \sum_i \b{Y}_{ij}\mbox{,}\\
{\nabla_{\b{G}_{nj}}^n} = &\sum_i \frac{\b{X}_{ij}\sum_{m}
    \b{P}_{im}\b{W}_{mn}}{\sum_{mn} \b{P}_{im}\b{W}_{mn}\b{H}_{nj}} +
 \sum_i \frac{\b{Y}_{ij}\sum_{m} \b{Q}_{im}\b{W}_{mn}}{\sum_{mn}
  \b{Q}_{im}\b{W}_{mn}\b{H}_{nj}}
\end{align*}
and
\[
\b{V}^{t+1} = \b{V}^t \odot \frac{\nabla_{\b{V}}^-}{\nabla_{\b{V}}^+}\mbox{,}
\]
with
\begin{align*}
{\nabla_{\b{V}_{mn}}^d} = &\sum_{ij} \frac{\b{X}_{ij}\sum_m
  \b{P}_{im}\b{W}_{mn}\b{H}_{nj}}{\sum_{mn}
  \b{P}_{im}\b{W}_{mn}\b{H}_{nj}} + 
\sum_{ij} \frac{\b{Y}_{ij}\sum_m
  \b{Q}_{im}\b{W}_{mn}\b{H}_{nj}}{\sum_{mn}
  \b{Q}_{im}\b{W}_{mn}\b{H}_{nj}}\mbox{,} \\ 
{\nabla_{\b{V}_{mn}}^n} = &\sum_{ij}
\frac{\b{X}_{ij}\b{P}_{im}\b{H}_{nj}}{\sum_{mn}
  \b{P}_{im}\b{W}_{mn}\b{H}_{nj}} +
 \sum_{ij} \frac{\b{Y}_{ij}\b{Q}_{im}\b{H}_{nj}}{\sum_{mn}
  \b{Q}_{im}\b{W}_{mn}\b{H}_{nj}}\mbox{.} 
\end{align*}

\tikzstyle{splate caption} = [caption, node distance=0, inner sep=0pt,
above left=40pt and 0pt of #1.south east] %
\newcommand{\splate}[4][]{ %
  \node[wrap=#3] (#2-wrap) {}; %
  \node[splate caption=#2-wrap] (#2-caption) {#4}; %
  \node[plate=(#2-wrap)(#2-caption), #1] (#2) {}; %
}

\begin{figure}[t]
 \centering
 \begin{tikzpicture}[font=\normalfont]
	\node[rectangle, draw=black](AA){\makecell{AA\\\cite{cutler_adele_archetypal_1994}}};
	\node[rectangle, draw=black, right=1.0cm of AA](CNMF){\makecell{C-NMF\\\cite{Ding:2010}}};
	\node[rectangle, draw=black, below=1.5cm of CNMF](PLSA){\makecell{PLSA\\\cite{hofmann_probabilistic_2013}}};
	\node[rectangle, draw=black, below=0.7cm of PLSA](NMF){\makecell{NMF\\\cite{lee_learning_1999}}};
	\node[rectangle, draw=black, below=1.5cm of AA](VCA){\makecell{VCA\\\cite{nascimento_vertex_2005}}};
	\node[rectangle, draw=black, below=0.7cm of VCA](SNMF){\makecell{S-NMF\\\cite{Ding:2010}}};
	\node[rectangle, draw=black, left=2.07cm of AA](PAA){\makecell{PAA\\}};
	\node[rectangle, draw=black, left=0.3cm of VCA](EFPCA){\makecell{EF-PCA\\\cite{DBLP:conf/nips/MohamedHG08}}};
	\node[rectangle, draw=black, below=1.0cm of EFPCA](PPCA){\makecell{PPCA\\}};
	\node[rectangle, right=0.5cm of CNMF](H1){};
	\node[rectangle, right=0.5cm of PLSA](H4){};
	\node[rectangle, below=0.5cm of SNMF](H2){};
	\node[rectangle, below=0.5cm of NMF](H3){};
	\node[rectangle, right=1.5cm of PPCA](H5){};
	
	\plate [very thick] {DD}{(PAA)(AA)(CNMF)(H1)}{1. $\b{Z} = \b{X}\b{W}$};
	\plate [very thick] {NN}{(NMF)(PLSA)}{5. $\b{Z},\b{H}$ nonnegative};
	\plate [very thick] {PP}{(PAA)(EFPCA)(PPCA)}{3. $\b{X} \sim f(\b{Z}\b{H})$};
	\plate [very thick] {MM}{(VCA)(SNMF)}{4. $\b{H}$ nonnegative};
	\splate [very thick] {SS}{(PLSA)(VCA)(H4)}{2. $\b{H}$ stochastic};

        \draw[dashed] (PAA) -- (EFPCA);
	\draw[dashed] (EFPCA) -- (PPCA);
	\draw[dashed] (SNMF) -- (NMF);
	\draw[dashed] (NMF.east) -- ++(0.8,0) |- (CNMF.east);
	\draw[dashed] (PLSA) -- (VCA);
	\draw[dashed] (AA) -- (CNMF);
	\draw[dashed] (VCA) -- (SNMF);
	\draw[dashed] (PAA.south) -- ++(0,-0.9) -| (VCA.north);
	\draw[dashed] (PAA.south) -- ++(0,-0.9) -| ([xshift=-0.5cm]PLSA.north);
	\draw[dashed] (AA) -- (PAA);
	\draw[dashed] (PLSA) -- (NMF);

  \end{tikzpicture}
  \caption{Relations among factorization methods. 1. data-driven methods
where the loadings depend on input, 2. simplex factor models with probability
vector as factors, 3. probabilistic methods, 4. non-negative factors with
arbitrary loadings, and 5. non-negative factors with non-negative loadings.
The connections are elaborated in section~\ref{sec:foundations}.}
  \label{fig:PAA}
\end{figure}
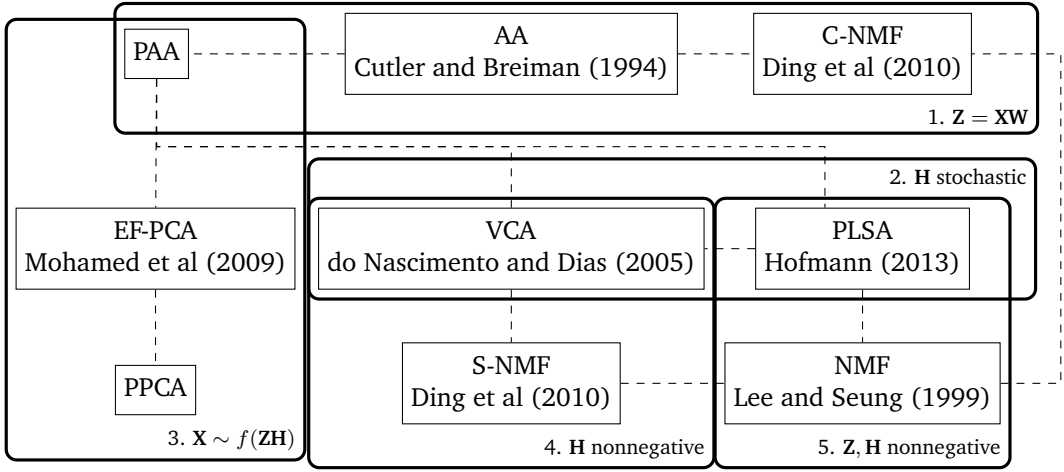

\section{Related work}
\label{sec:foundations}

Archetypal analysis and its probabilistic extension share close 
connections with other popular matrix factorization methods.
We explore some of these connections in this section, and provide
a summary in Figure~\ref{fig:PAA}. We represent the original data
matrix by $\b{X} \in \mathcal{X}^{M \times N}$ where each column
$\b{x}_n$ is an observation; the corresponding latent factor matrix by
$\b{H} \in \reals^{K \times N}$, and loading matrix by $\b{Z} \in
\reals^{M\times K}$.

\paragraph{Principal component analysis and extensions:}
Principal component analysis (PCA) finds an orthogonal transformation
of a point-cloud, which projects the observations in a new coordinate
system that preserves the variance of the point-cloud the best. The
concept of PCA has been extended to a probabilistic as well as a
Bayesian framework \citep{DBLP:conf/nips/MohamedHG08}. Probabilistic
PCA (PPCA) assumes that the data originates from a lower dimensional
subspace on which it follows a normal distribution ($\normal$), i.e., 
\[
\b{h}_n \sim \normal(0,\b{I}),\,
\b{x}_n \sim \normal(\b{Z}\b{h}_n,\epsilon\b{I})
\]
where $\b{h}_n \in \reals^K$, $\b{x}_n \in \reals^M$, $K < M$, and
$\epsilon > 0$.

Probabilistic principal component analysis explicitly assumes that the
observations are \emph{normally distributed}: an assumption that is
often violated in practice, and to tackle such situations one extends
PPCA to exponential family ($\EF$). The underlying principle here is
to change the observation model accordingly: 
\[ \b{h}_n \sim \normal(0,\b{I}), \,\b{x}_n \sim \EF(\b{x}_n;\b{Z}\b{h}_n),\]
 i.e.,
each element of $\b{x}_n$ is generated from the corresponding element
of $\b{Z}\b{h}_n$ as $\EF(z;\theta) = h(z)g(\theta)\exp(\theta s(z))$
where $s(z)$ is the sufficient statistic, and $\theta$ is the
\emph{natural parameter}: PAA utilizes similar approach but with
normal parameters.

Similarly, one can also manipulate the latent distribution. A popular
choice is the Dirichlet distribution ($\Dir$), which has been widely
explored in the literature, e.g., in probabilistic latent semantic
analysis (PLSA~\citep{hofmann_probabilistic_2013}), $\b{h}_n \sim
\Dir(\b{1}), \,\b{x}_n \sim \Mult(\b{Z}\b{h}_n), \mbox{ where }
\b{1}\b{Z} = \b{1}$; vertex component analysis
\citep{nascimento_vertex_2005}, $\b{h}_n \sim \Dir(\b{1}), \,\b{x}_n
\sim \normal(\b{Z}\b{h}_n,\epsilon\b{I})$; and simplex factor analysis
\citep{bhattacharya_simplex_2012}, a generalization of PLSA (or more
specifically of latent Dirichlet allocation, LDA~\citep{Blei:2003:LDA}):
PAA additionally decompose the loading in simplex factors with known loading.

\begin{figure*}[t]
\centering
\includegraphics[width=\textwidth]{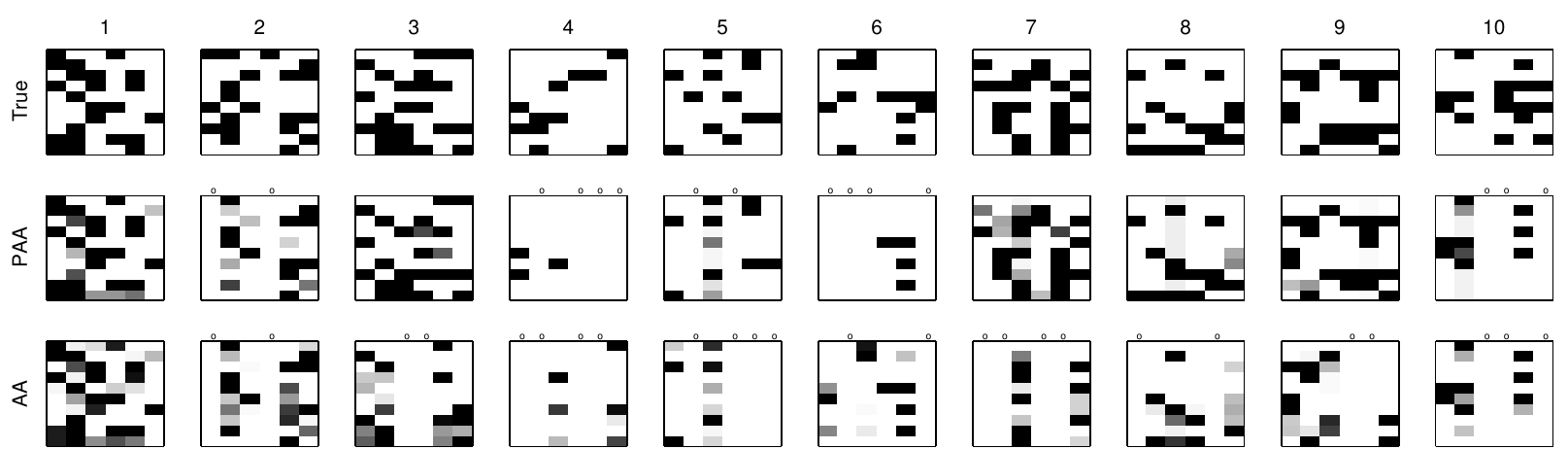}
\caption{The figure compares the solutions achieved by standard archetypal
analysis and the probabilistic formulation on binary observations. 
 Each column is an independent trial, and each algorithm has been run 10 times to 
find the best archetypal profiles. The
archetypal profiles are binarized, and matched with the true archetypes using
minimum Jaccard distance. If a unique match is found then the 
corresponding archetypal profile is displayed. Otherwise they are left blank, 
and tagged by a circle. We
observe that the probabilistic approach has been able to match archetypes 
better than the standard solution.
}
\label{fig:compareProbVsStandardBinary}
\end{figure*}

\paragraph{Nonnegative matrix factorization and extensions:}
Non-negative matrix factorization (NMF) decomposes a non-negative
matrix $\b{X} \in \reals_+^{M\times N}$ in two non-negative matrices
$\b{Z} \in \reals_+^{M\times K}$ and $\b{H} \in\reals_+^{K\times N}$
such that $\b{X} \approx \b{Z}\b{H}$. \citep{lee_learning_1999}
applied the celebrated \emph{multiplicative update rule} to solve this
problem, and proved that such update rules lead to monotonic decrease
in the cost function using the concept of
\emph{majorization-minimization} 
\citep{lee_algorithms_2000}.  Non-negative matrix factorization has
been extended to convex non-negative matrix factorization
(C-NMF~\citep{Ding:2010}) where $\b{X}$ is not restricted to be
non-negative, and $\b{Z}$ is expressed in terms of the $\b{X}$ itself
as $\b{Z} = \b{X}\b{W}$, where $\b{W}$ is again a non-negative
matrix. The motivation for this modification emerges from its
similarity to clustering, and C-NMF has been solved using
multiplicative update rule as well.

To simulate the exact clustering scenario, however, $\b{H}$ is
required to be binary (hard clustering) or at least column stochastic
(fuzzy clustering). This leads to a more difficult optimization
problem, and is usually solved by proxy constraint $\b{H}^\top\b{H} =
\b{I}$~\citep{ding_orthogonal_2006}. Several other alternatives have
also been proposed for tackling the stochasticity constraints, e.g.,
by enforcing it after each iteration \citep{morup_archetypal_2012}, or
by employing a gradient-dependent Lagrangian multiplier
\citep{yang_clustering_2012}. However, both these approaches are prone to
finding local minima.

\begin{figure*}[t]
\centering
\includegraphics[width=\textwidth]{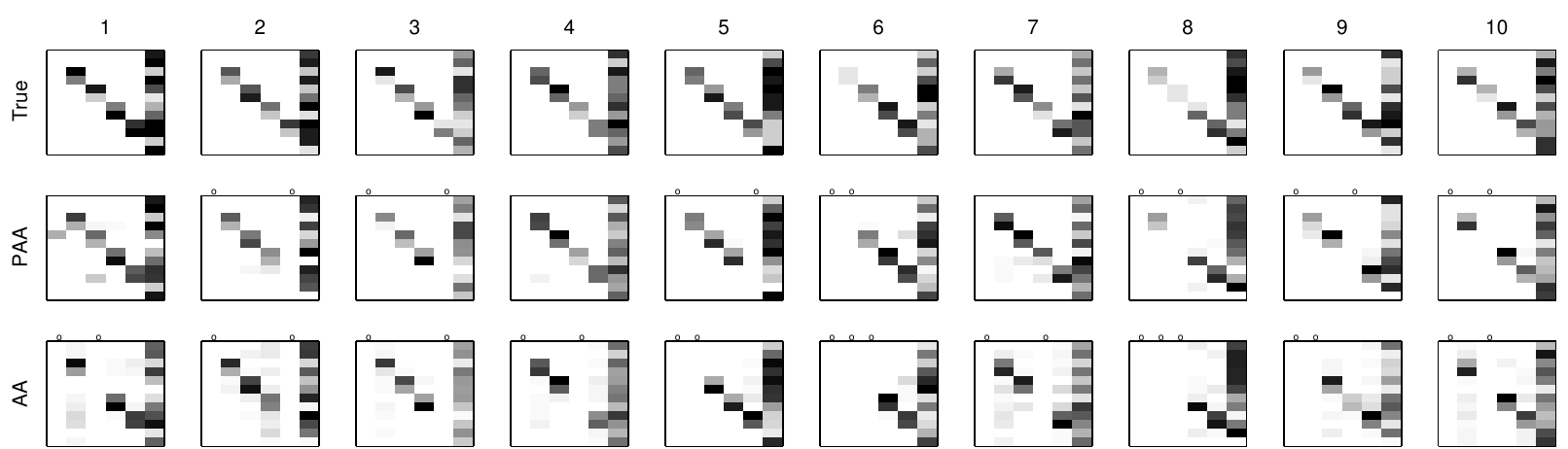}
\caption{The figure compares the solutions achieved by standard archetypal
analysis and the probabilistic formulation on count observations. 
Each column is an independent trial, and each algorithm has been run 10 times
to find the best archetypal profiles. The archetypal profiles are matched with
the true archetypes using minimum $l_1$ distance. If a unique match is found
then the corresponding archetypal profile is displayed. Otherwise they are left
blank, and tagged by a circle. We observe that the probabilistic approach has
been able to match archetypes better than the standard solution.
}
\label{fig:compareProbVsStandardPoisson}
\end{figure*}

\section{Simulation}
\label{sec:simulation}

In this section, we provide some simple examples showing the difference
between probabilistic and standard archetypal analysis solutions. Since we
generate data following the true probabilistic model, it is expected that the 
solution provided by PAA would be more appropriate compared to the standard
AA solution. Therefore, the purpose of this section is 
to perform sanity check, and provide insight. Notice that generating 
observations with known archetypes is not straight forward, since $\bs{\Theta}$
depends on $\b{X}$.

\paragraph{Binary observations:} 
We generate $K=6$ binary archetypes in $d=10$ dimensions by sampling
$\eta_{ik} \sim \mathrm{Bernoulli}(p_s)$, where $\bs{\eta}_{k}$ is an
archetype, and $p_s=0.3$ is the probability of success. Given the 
archetypes, we generate $n=100$ observations as 
$\b{x}_i \sim \mathrm{Bernoulli}(\b{E}\b{h}_i)$, where 
$\b{E}=[\bs{\eta}_1,\ldots,\bs{\eta}_k]$, and each $\b{h}_i$ is a stochastic vector sampled from $\Dir(\alpha)$. To
ensure that $\bs{\eta}$'s are archetypes, we maintain more observations around
$\bs{\eta}_k$s by choosing $\alpha=0.4$. We find archetypal profiles using both PAA
and standard AA, and then binarize them so that they
can be matched to the original archetypes 
using minimum Jaccard distance. 
We report the results in Fig.~\ref{fig:compareProbVsStandardBinary}. 
We observe that PAA has been more successful in finding the
true archetypes.
\begin{figure*}[t]
\centering
\includegraphics[width=\textwidth]{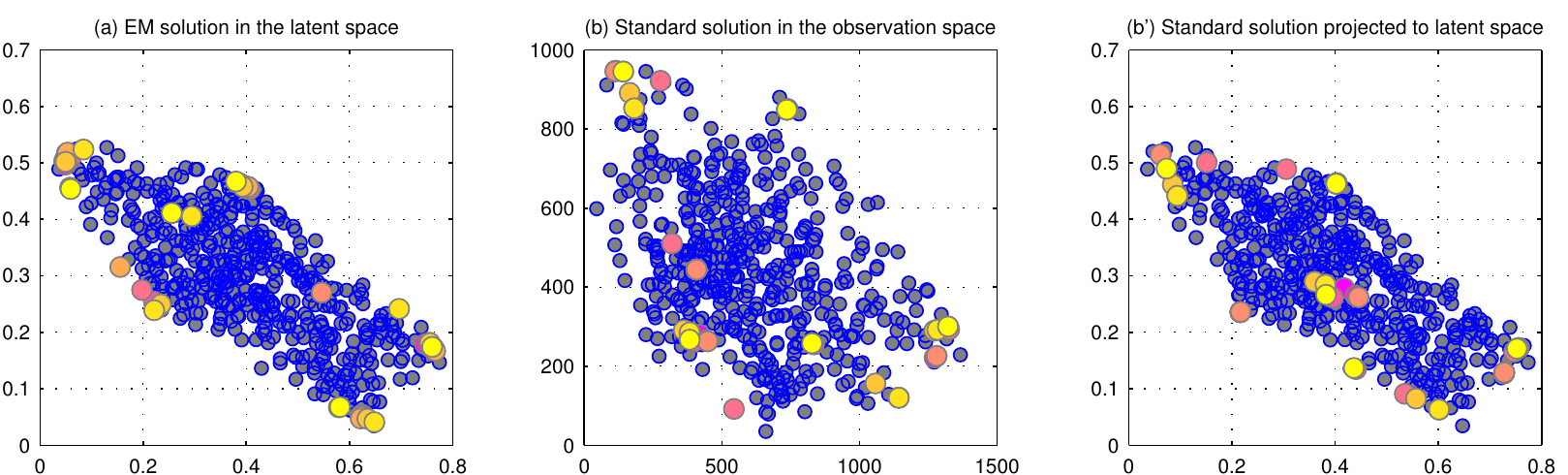}
\caption{The figure compares the solutions achieved by standard archetypal analysis
and probabilistic archetypal analysis on term-frequency observations. 
Each observation vector of the term-frequency matrix is generated
from a probability vector within a clear convex hull (a). The probability
vectors are generated such there are 5 archetypes. However, this structure is lost
in the term-frequency values due to arbitrary number of occurrences in each
term-frequency vector (b). The standard AA applied to term-frequency matrix
thus, does not capture the true archetypes (b').}
\label{fig:compareProbVsStandard}
\end{figure*}

\paragraph{Poisson observations:}
We generate $K=6$ count archetypes in $d=12$ dimensions with one minimal
archetype ($\eta_{ik}=0$), one maximal archetype $\eta_{ik} \sim \mathrm{Unif}
\{1,\ldots,10\}$, and rest of the archetpyes with two nonzero entries 
$\eta_{ik}\ \sim \mathrm{Unif}\{1,\ldots,10\}$. Given the 
archetypes, we generate $n=500$ observations as 
$\b{x}_i \sim \mathrm{Poisson}(\b{E}\b{h}_i)$, where 
$\b{E}=[\bs{\eta}_1,\ldots,\bs{\eta}_k]$, and each $\b{h}_i$ is a stochastic vector sampled from $\Dir(\alpha)$. To
ensure that $\bs{\eta}$'s are archetypes, we maintain more observations around
$\bs{\eta}_k$s by choosing $\alpha=0.4$. We find archetypal profiles using both PAA
and standard AA, and match them to the original archetypes 
using minimum $l_1$ distance. 
We report the results in Fig.~\ref{fig:compareProbVsStandardPoisson}. 
We observe that PAA has been more successful in finding the
true archetypes.

\paragraph{Term-frequency observations:}
We generate $K=5$ archetypes on $d=3$ dimensional probability simplex by
choosing $K$ equidistant points $\b{p}_k$ on a circle in the simplex. Given the 
archetypes, we generate $n=500$ observations as 
$\b{x}_i \sim \mathrm{Mult}(n_i,\b{P}\b{h}_i)$, where 
$\b{P}=[\bs{p}_1,\ldots,\bs{p}_k]$, and each $\b{h}_i$ is a stochastic vector sampled from $\Dir(\alpha)$. To
ensure that $\b{p}$'s are archetypes, we maintain more observations around
them by choosing $\alpha=0.5$. 
We deliberately choose an arbitrary number of occurrences $n_i \sim \mathrm{Uniform}
[1000,2000]$ for each observation:
this disrupts the true convex hull structure in the term-frequency observations. We
present 10 random runs on these observations for both PAA and standard AA in
Fig.~\ref{fig:compareProbVsStandard}. We observe that  PAA finds the 
effective archetypes, with occasional local minima. However, standard 
AA performs poorly since it finds the appropriate archetypes in the term-frequency
space, which are different when projected back on the simplex.

\section{Simplex visualizations}\label{sec:vis}

\begin{figure}[p]
  \centering
  \subfigure[]{\includegraphics[scale=0.8]{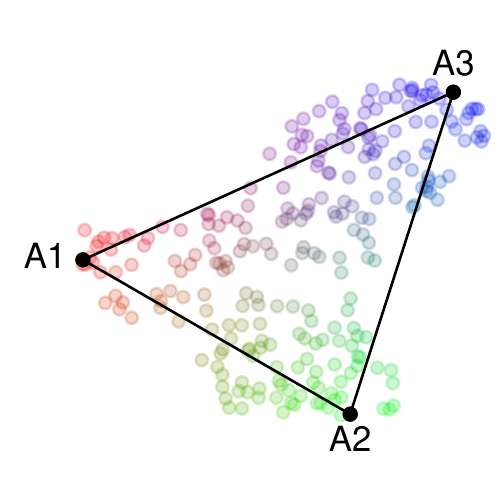}}%
  \subfigure[]{\includegraphics[scale=0.8]{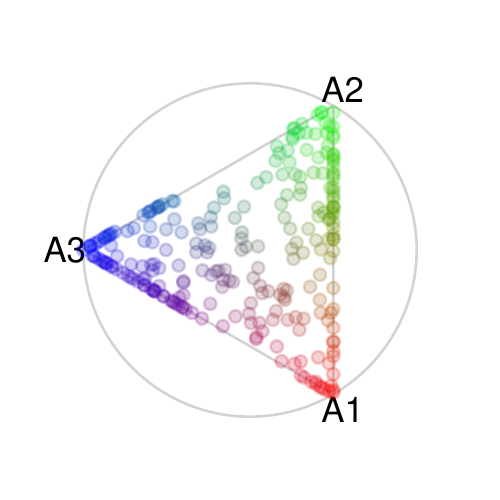}}%
  \subfigure[]{\includegraphics[scale=0.8]{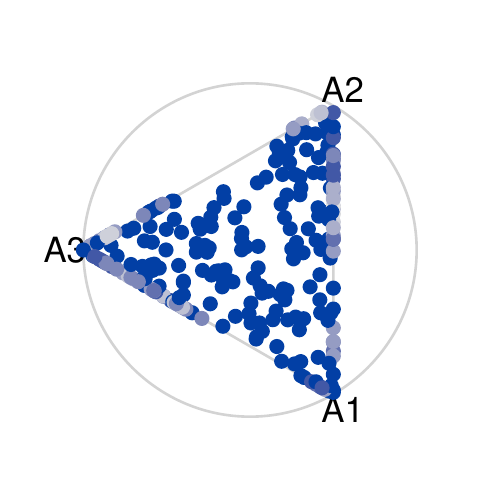}}\\
  \subfigure[]{\includegraphics[scale=0.8]{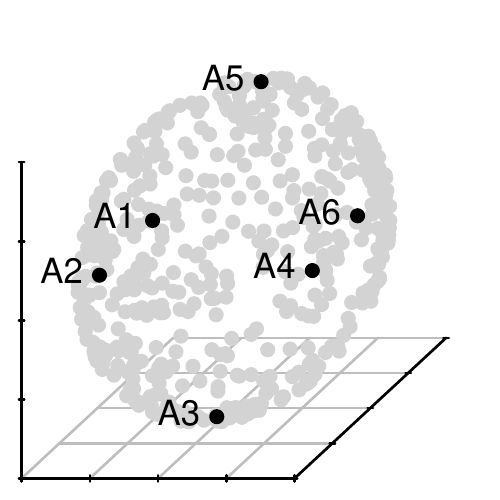}}%
  \subfigure[]{\includegraphics[scale=0.8]{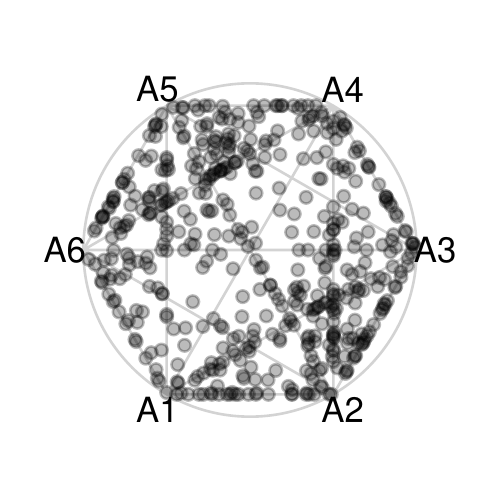}}%
  \subfigure[]{\includegraphics[scale=0.8]{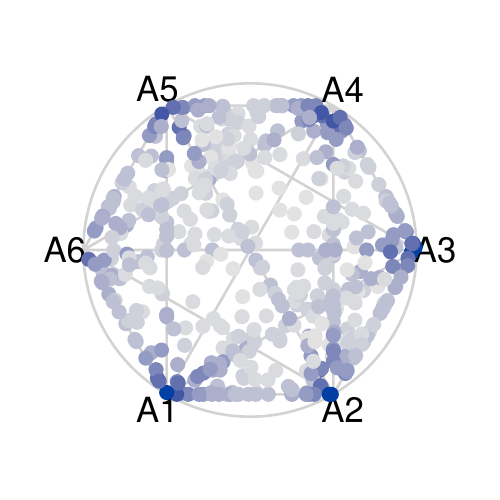}}\\
  \subfigure[]{\includegraphics[scale=0.8]{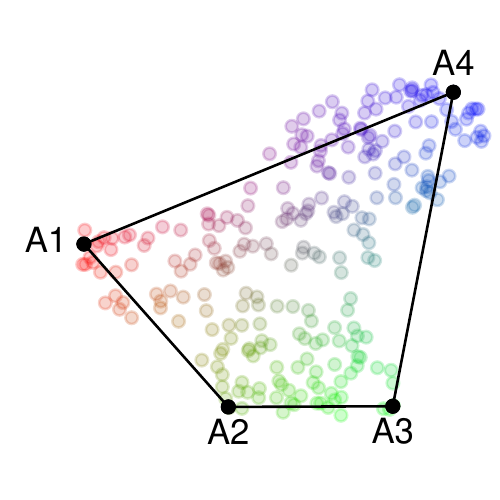}}%
  \subfigure[]{\includegraphics[scale=0.8]{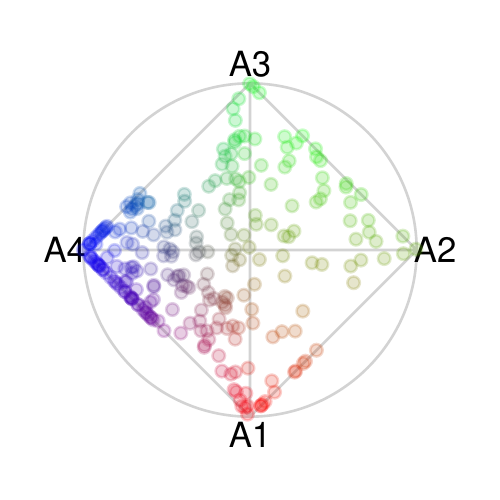}}%
  \subfigure[]{\includegraphics[scale=0.8]{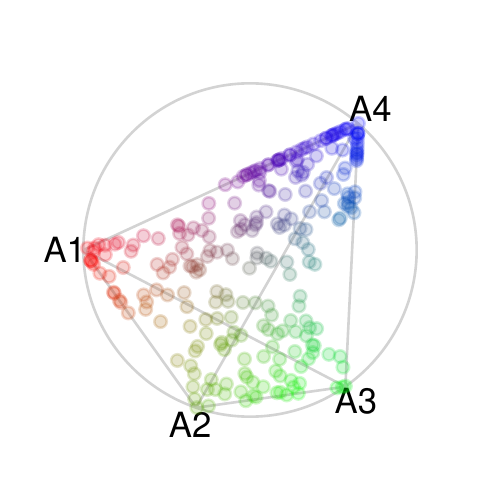}}\\
  \subfigure[]{\includegraphics[scale=0.8]{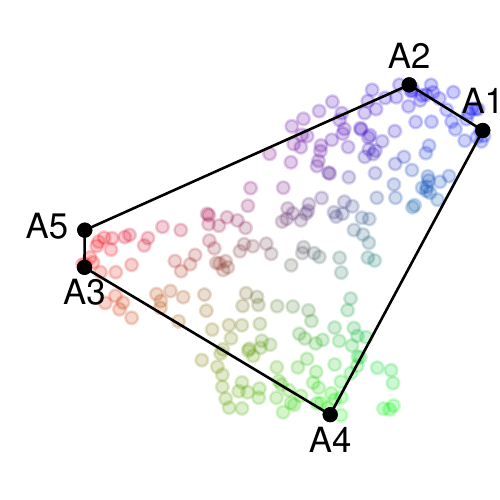}}%
  \subfigure[]{\includegraphics[scale=0.8]{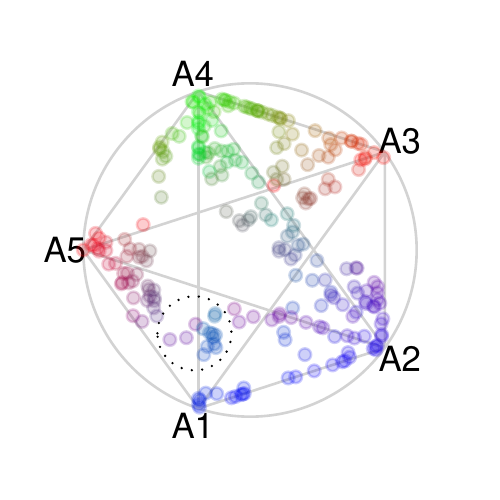}}%
  \subfigure[]{\includegraphics[scale=0.8]{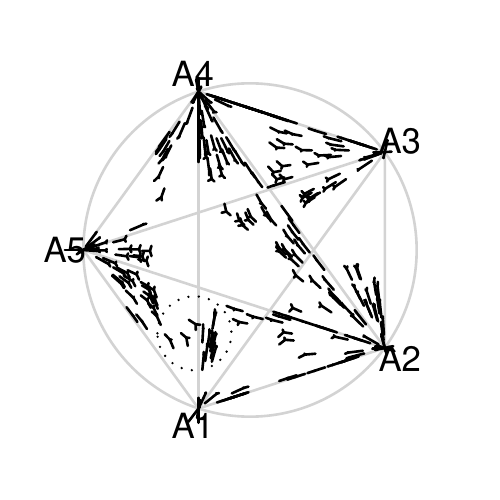}}
  \caption{Simplex visualizations with extensions for different
    illustrative data sets and AA solutions: figures~(a-c) and (d-f)
    illustrate color-coded points based on the deviance; figures~(g-i)
    illustrate the ordering of the vertices according to the distances
    of the archetypes in the original space; and figures~(j-l)
    illustrate the enhancement of the plot to show the composition of
    observations.}
  \label{fig:simplex}
\end{figure}

The column stochasticity of $\b{H}$ allows a principled visualization
scheme of archetypal analysis solution, referred to as \emph{simplex
visualization}. We discuss certain aspects and enhancements of this 
approach, and show how it can be utilized to better understand the
inferred archetypes. 

The stochastic
nature of $\b{h}_n$ implies that $\b{Z}\b{h}_n$ exists on a standard
$(K-1)$-simplex with the $K$ archetypes $\b{Z}$ as the corners, and
$\b{h}_n$ as the coordinate with respect to these corners (see
Figure~\ref{fig:AA}b for an illustration). A standard simplex can be
projected to two dimensions via a skew orthogonal projection, where
all the vertices of the simplex are shown on a circle connected by
edges. The individual factors $\b{h}_n$ can be then projected into
this circle.  Figure~\ref{fig:simplex} illustrates this principle with
the simple data set already used in Figure~\ref{fig:AA}: (a,~b)~for
the three archetypes solution; (g,~h)~for the four archetypes
solution; (j,~k)~for the five archetypes solution. Color coding is
used in the six figures to show the relation between the original
observation $\b{x}_n$ and its projection $\b{h}_n$. The visualization
with three archetypes is known as ternary plot \citep[see,
  e.g.,][]{Friendly:2000}, and has been used by
\citet{cutler_adele_archetypal_1994}. The extension to more than three
archetypes has also been used \citep[e.g.,][]{Bauckhage+Thurau@2009,
  archetypes}. However, a formal study of this visualization scheme,
to the best of our knowledge, remains to be explored.
In the following, we present three enhancements of this basic
visualization to either highlight certain characteristics of an
archetypal analysis solution or to overcome consequences of the
one-to-many projection.

In AA, observations which lie outside the approximated convex hull are
projected onto its boundary. Figures~\ref{fig:simplex}a and b show a
simple scenario where these observations are projected onto the
corresponding edges. Figures~\ref{fig:simplex}d and e show an
extreme case of this characteristic: the observations lie on a
three-dimensional sphere, and therefore the computed archetypes span
a space, which is completely empty. In the corresponding simplex
visualization, however, this aspect of the solution is not visible
at all. We propose to visualize the `deviance' 
$D(\b{x}_n) = 2(\log{p(\b{x}_n|\theta_n)} - \log{p(\b{x}_n|\b{Z}\b{h}_n)})$
where $\theta_n$ is the maximum likelihood estimate of $\b{x}_n$,
as colors of the points. 
In case of normal observation model the deviance
reduces to the residual sum of squares. Figure~\ref{fig:simplex}c
shows the corresponding simplex visualization with deviance
for the three archetypes solution. The color scheme is from
blue to white, with blue implying zero deviance and the lighter the color,
the higher the deviance. 
We can now identify how well the original observations have been
approximated by the archetypes, and if they are inside the convex hull.
This extension is even more insightful
in case of the sphere example. In the corresponding simplex
visualization in Figure~\ref{fig:simplex}f, we can now clearly see
that almost all observations are outside the approximated convex hull;
only around the corners the deviance is near to zero.

The basic simplex visualization arranges the vertices, which
represents the archetypes, equidistant on the circle. The archetypes
in the original space, however, are usually not equidistant to each
other. Figure~\ref{fig:simplex}g and h illustrate this discrepancy:
archetype A2, for example, is much nearer to A3 than A4; in the
simplex visualization, however, both are in the same distance. We propose to
order the vertices on the circle according to their distances in the
original space. This means, we first have to determine an optimal
order of the vertices, and then divide the 360$^\circ$ of the circle
in relation to the original pairwise distances of the determined
neighbor vertices. Here, we solve a Traveling Salesman Problem to get
an optimal cyclic order \citep[solved by using, for
  example,][]{Hahsler+Hornik:2007}; and then simply divide the circle
proportional to the original distances. Figure~\ref{fig:simplex}i
shows the result: it is now clearly visible that A2 and A3 are much
nearer to each other than A3 and A4.

Another problem of the simplex visualization with more than three
archetypes is the \emph{non-uniqueness}. As a result two projections 
$\b{h}_{n1}$ and $\b{h}_{n2}$ can be close to each other even though
they are \emph{composed} of different archetypes. This goes against 
one's intuition in judging which archetypes the observations belong to.
For example, the observations inside the
dashed circle in Figure~\ref{fig:simplex}k. We get the idea that
these observations are basically composed by A1, A2, A5, and/or
A4---but we do not get the exact compositions. We therefore propose to
show `whiskers', which point in the direction of the composing
archetypes, Figure~\ref{fig:simplex}l shows the corresponding
visualization. We can now easily see the composition of the
observations inside the dashed circle: the observations on the right
side of the line are composed by A1 and A4; the observations on the
left side of the line are composed by A1, A2, A5; and the left-most
observation is composed by A1, A5. We vary the length of the
`whiskers' according to the coefficients $\b{h}_n$; the longer the
whisker the closer the observation is to the archetype the whisker
points toward.

\begin{figure*}[t]
  \centering
  \includegraphics[width=\textwidth]{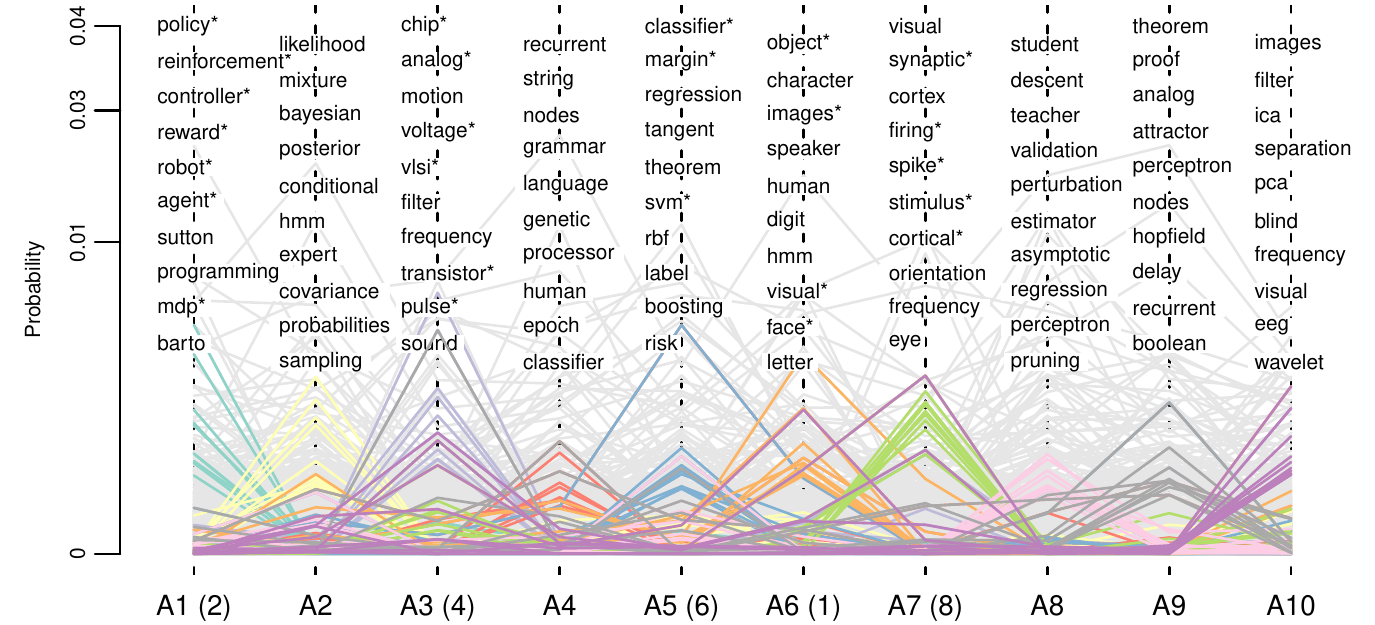}
  \caption{The probability of the words available in the NIPS corpus
    for each of the ten archetypal profiles. The number in parentheses
    refers to the corresponding archetype in
    \citep{morup_archetypal_2012}. The colored lines show the ten most
    prominent words (after removing the ``common'' words),
    \mo~indicates which words appear in both solutions. More
    information in Section~\ref{subsec:nips}.}
  \label{fig:ex-nips}
\end{figure*}

\section{Applications}
\label{sec:experiments}

In each application below (except the first one), we run PAA with
$2\mbox{--}15$ archetypes, $10$ trials with random initializations for
each number of archetypes, and choose the solution from the trials
with maximum likelihoods according to the ``elbow criterion''.
The elbow criterion is a simple heuristic. Due to the fact that with
each additional archetype $\mathbb{LL}$ increases, one can compute
solutions with successively increasing number of archetypes, plot
$\mathbb{LL}$ against the number or archetypes, and visually pick the
solution after which the jump of $\mathbb{LL}$ ``is only marginal''
(i.e., the elbow). This solution is ad hoc and subjective---but widely
used: ``Statistical folklore has it that the location of such an
'elbow' indicates the appropriate number of clusters''
\citep{Tibshirani+Walther@2005}.

\subsection{Multinomial observations: NIPS bag-of-words} 
\label{subsec:nips}

We use a data set already explored by \citep{morup_archetypal_2012}
for archetypal analysis, to qualitatively 
evaluate the solution provided by PAA with multinomial observation
model. We analyze the NIPS bag-of-words corpus consisting of $N =
1500$ documents and $M = 12419$ words (available
from \citep{Bache+Lichman:2013}) and compute $K = 10$ document archetypes,
as in \citep{morup_archetypal_2012}. We use the term-frequencies as
features without normalizing them by the document frequency as in
\citep{morup_archetypal_2012} to adhere to the generative nature of the
documents. Figure~\ref{fig:ex-nips} shows the probability for each
word available in the corpus to be generated by the corresponding
archetype ($\b{Z}$). Following \citep{morup_archetypal_2012}, we
highlight the ten most prominent terms after ignoring the ``common''
terms, which are present in each of the archetypes in the first 3000
words (with probability values $>10^{-4}$). We can observe that the
prominent terms in a particular archetype have low probability in all
the other archetypes: this agrees with our understanding of an
archetype. Overall our algorithm finds a similar solution to
\citet{morup_archetypal_2012}---one difference, however, protrudes: we
find a ``Bayesian Paradigm'' archetype~(A2), which
\citet{morup_archetypal_2012} finds as a $k$-means prototype. But, to
the best of our knowledge a ``Bayesian Paradigm'' archetypal document
can make sense in a NIPS corpus.

\subsection{Bernoulli observations: Austrian national guest survey}
\label{subsec:tourist}

\begin{table*}[th]
  \centering
{
\begin{tabular}{r|cccccc}
  \hline
 & A1 & A3 & A5 & A4 & A6 & A2 \\ 
  \hline
Alpine Ski & \textcolor{black}{1.00 (1)} & \textcolor{black}{1.00 (1)} & \textcolor{black}{1.00 (1)} & \textcolor{black}{1.00 (1)} & \textcolor{gray}{0.00 (0)} & \textcolor{gray}{0.00 (0)} \\ 
  Tour Ski & \textcolor{black}{0.41 (1)} & \textcolor{gray}{0.00 (0)} & \textcolor{gray}{0.00 (0)} & \textcolor{gray}{0.00 (0)} & \textcolor{gray}{0.00 (0)} & \textcolor{gray}{0.00 (0)} \\ 
  Snowboard & \textcolor{gray}{0.00 (0)} & \textcolor{gray}{0.00 (0)} & \textcolor{gray}{0.00 (0)} & \textcolor{black}{0.59 (1)} & \textcolor{gray}{0.00 (0)} & \textcolor{gray}{0.00 (0)} \\ 
  Cross Country & \textcolor{black}{0.75 (1)} & \textcolor{gray}{0.00 (0)} & \textcolor{gray}{0.00 (0)} & \textcolor{gray}{0.00 (0)} & \textcolor{gray}{0.00 (0)} & \textcolor{gray}{0.00 (0)} \\ 
  Ice Skating & \textcolor{black}{0.60 (1)} & \textcolor{gray}{0.00 (0)} & \textcolor{gray}{0.00 (0)} & \textcolor{gray}{0.00 (0)} & \textcolor{gray}{0.00 (0)} & \textcolor{gray}{0.00 (0)} \\ 
  Sledge & \textcolor{black}{1.00 (1)} & \textcolor{gray}{0.00 (0)} & \textcolor{gray}{0.00 (0)} & \textcolor{gray}{0.00 (0)} & \textcolor{gray}{0.00 (0)} & \textcolor{gray}{0.00 (0)} \\ 
  Tennis & \textcolor{black}{0.15 (0)} & \textcolor{gray}{0.00 (0)} & \textcolor{gray}{0.00 (0)} & \textcolor{gray}{0.00 (0)} & \textcolor{gray}{0.00 (0)} & \textcolor{black}{0.20 (0)} \\ 
  Riding & \textcolor{gray}{0.00 (0)} & \textcolor{gray}{0.00 (0)} & \textcolor{gray}{0.00 (0)} & \textcolor{gray}{0.00 (0)} & \textcolor{gray}{0.00 (0)} & \textcolor{black}{0.08 (0)} \\ 
  Pool Sauna & \textcolor{black}{0.96 (1)} & \textcolor{gray}{0.00 (0)} & \textcolor{black}{0.37 (0)} & \textcolor{black}{1.00 (1)} & \textcolor{black}{0.82 (1)} & \textcolor{black}{0.11 (0)} \\ 
  Spa & \textcolor{black}{0.22 (0)} & \textcolor{gray}{0.00 (0)} & \textcolor{gray}{0.00 (0)} & \textcolor{gray}{0.00 (0)} & \textcolor{black}{0.79 (1)} & \textcolor{gray}{0.00 (0)} \\ 
  Hiking & \textcolor{black}{0.95 (1)} & \textcolor{gray}{0.00 (0)} & \textcolor{gray}{0.00 (0)} & \textcolor{gray}{0.00 (0)} & \textcolor{black}{1.00 (1)} & \textcolor{black}{0.18 (0)} \\ 
  Walk & \textcolor{black}{1.00 (1)} & \textcolor{gray}{0.00 (0)} & \textcolor{black}{1.00 (1)} & \textcolor{gray}{0.00 (0)} & \textcolor{black}{1.00 (1)} & \textcolor{black}{1.00 (1)} \\ 
  Excursion (org) & \textcolor{black}{0.29 (0)} & \textcolor{gray}{0.00 (0)} & \textcolor{gray}{0.00 (0)} & \textcolor{gray}{0.00 (0)} & \textcolor{gray}{0.00 (0)} & \textcolor{black}{0.41 (1)} \\ 
  Excursion (ind) & \textcolor{black}{0.81 (1)} & \textcolor{gray}{0.00 (0)} & \textcolor{gray}{0.00 (0)} & \textcolor{gray}{0.00 (0)} & \textcolor{black}{0.94 (1)} & \textcolor{black}{1.00 (1)} \\ 
  Relax & \textcolor{black}{0.99 (1)} & \textcolor{gray}{0.00 (0)} & \textcolor{black}{1.00 (1)} & \textcolor{black}{1.00 (1)} & \textcolor{black}{1.00 (1)} & \textcolor{black}{0.81 (0)} \\ 
  Dinner & \textcolor{black}{0.82 (1)} & \textcolor{black}{0.53 (0)} & \textcolor{black}{0.86 (1)} & \textcolor{black}{0.02 (0)} & \textcolor{gray}{0.00 (0)} & \textcolor{black}{1.00 (1)} \\ 
  Shopping & \textcolor{black}{1.00 (1)} & \textcolor{gray}{0.00 (0)} & \textcolor{black}{1.00 (1)} & \textcolor{black}{0.01 (0)} & \textcolor{black}{0.33 (0)} & \textcolor{black}{1.00 (1)} \\ 
  Concert & \textcolor{gray}{0.00 (0)} & \textcolor{gray}{0.00 (0)} & \textcolor{gray}{0.00 (0)} & \textcolor{gray}{0.00 (0)} & \textcolor{gray}{0.00 (0)} & \textcolor{black}{0.29 (1)} \\ 
  Sightseeing & \textcolor{black}{0.66 (1)} & \textcolor{gray}{0.00 (0)} & \textcolor{gray}{0.00 (0)} & \textcolor{gray}{0.00 (0)} & \textcolor{black}{0.66 (1)} & \textcolor{black}{1.00 (1)} \\ 
  Heimat & \textcolor{black}{0.58 (1)} & \textcolor{gray}{0.00 (0)} & \textcolor{gray}{0.00 (0)} & \textcolor{gray}{0.00 (0)} & \textcolor{gray}{0.00 (0)} & \textcolor{gray}{0.00 (0)} \\ 
  Museum & \textcolor{gray}{0.00 (0)} & \textcolor{gray}{0.00 (0)} & \textcolor{gray}{0.00 (0)} & \textcolor{gray}{0.00 (0)} & \textcolor{gray}{0.00 (0)} & \textcolor{black}{1.00 (1)} \\ 
  Theater & \textcolor{gray}{0.00 (0)} & \textcolor{gray}{0.00 (0)} & \textcolor{gray}{0.00 (0)} & \textcolor{gray}{0.00 (0)} & \textcolor{gray}{0.00 (0)} & \textcolor{black}{0.30 (1)} \\ 
  Heurigen & \textcolor{gray}{0.00 (0)} & \textcolor{gray}{0.00 (0)} & \textcolor{gray}{0.00 (0)} & \textcolor{gray}{0.00 (0)} & \textcolor{gray}{0.00 (0)} & \textcolor{black}{0.45 (0)} \\ 
  Local Event & \textcolor{black}{0.99 (1)} & \textcolor{gray}{0.00 (0)} & \textcolor{gray}{0.00 (0)} & \textcolor{gray}{0.00 (0)} & \textcolor{gray}{0.00 (0)} & \textcolor{black}{0.10 (0)} \\ 
  Disco & \textcolor{black}{0.68 (1)} & \textcolor{gray}{0.00 (0)} & \textcolor{gray}{0.00 (0)} & \textcolor{black}{1.00 (1)} & \textcolor{gray}{0.00 (0)} & \textcolor{black}{0.08 (0)} \\ 
   \hline
\hline
Interpretation & Maximal & Minimal & \multicolumn{2}{c}{Basic} &
\multicolumn{2}{c}{Non-Sportive} \\
  of the archetypes &  &  & Traditional & Modern & Wellness & Cultural \\
   \hline
\end{tabular}
}
  \caption{The six archetypal profiles for the winter tourists example. The corresponding archetypal observations are shown in parentheses. For more information, see Section~\ref{subsec:tourist}.}
  \label{tab:gsaw97-atypes}
\end{table*}

Analyzing binary survey data is of utmost importance in social science
and marketing research.  A binary questionnaire is often preferred
over an ordinal multi-category format, since the former is quicker and
easier, whereas both are equally reliable, and the managerial
implications derived from them do not substantially differ
\citep{Dolnicar+Grun+Leisch@2011}. In this application, we analyze
binary survey data from the Austrian National Guest Survey conducted
in the winter season of 1997. The goal is to identify archetypal
winter tourists, which may facilitate developing and targeting
specific advertising materials.  The data consists of $2958$
tourists. Each tourist answers $25$ binary questions on whether he/she
is engaged in a certain winter activity (e.g., alpine skiing,
relaxing, or shopping; see row description of
Table~\ref{tab:gsaw97-atypes} for the complete list). In addition, a
number of descriptive variables are available (e.g., the age and
gender of the tourist).

Here we present the six archetypes
solution. Table~\ref{tab:gsaw97-atypes} lists the archetypal profiles
(i.e., the probability of positive response) and, in parentheses, the
corresponding archetypal observations (with maximum $w$ value).
Archetype A1 is the maximal winter tourist who is engaged in nearly
every sportive and wellness activity with high probability. Archetype
A3, on the other hand, is the minimal winter tourist who is only
engaged in alpine skiing and having dinner. Both archetypes A5 and A4
are engaged in the basic Austrian winter activities (alpine skiing,
indoor swimming, and relaxing). In addition, A5 is engaged in
traditional activities (dinner and shopping), whereas A4 is engaged in
more modern activities (snowboarding and going to a disco). Finally,
A6 and A2 are the non-sportive archetypes. A6 is engaged in wellness
activities and A2 with cultural activities. Note that important
engagements of the archetypes are missed if one only looks at the
archetypal observations rather than the archetypal profiles; e.g., the
possible engagement of A2 in hiking. We can now utilize the factors
$\b{H}$ for each of the tourists to learn their relations to the
archetypal winter tourist profiles. This allows us, for example, to
target very specific advertising material to tourists for the next
winter season.

\begin{figure*}[h]
\centering
  \subfigure[]{\includegraphics[scale=0.5]{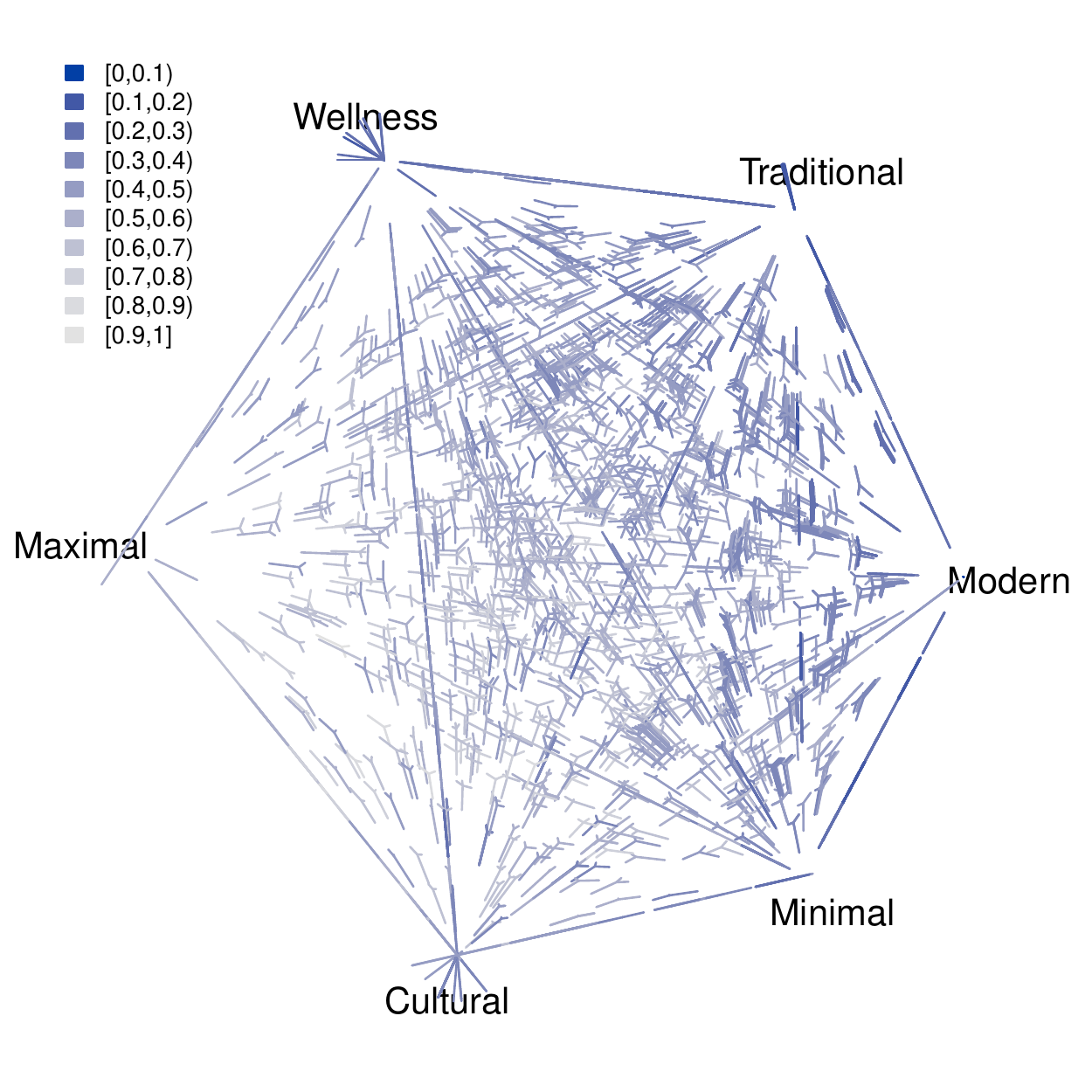}}
  \subfigure[]{\includegraphics[scale=0.5]{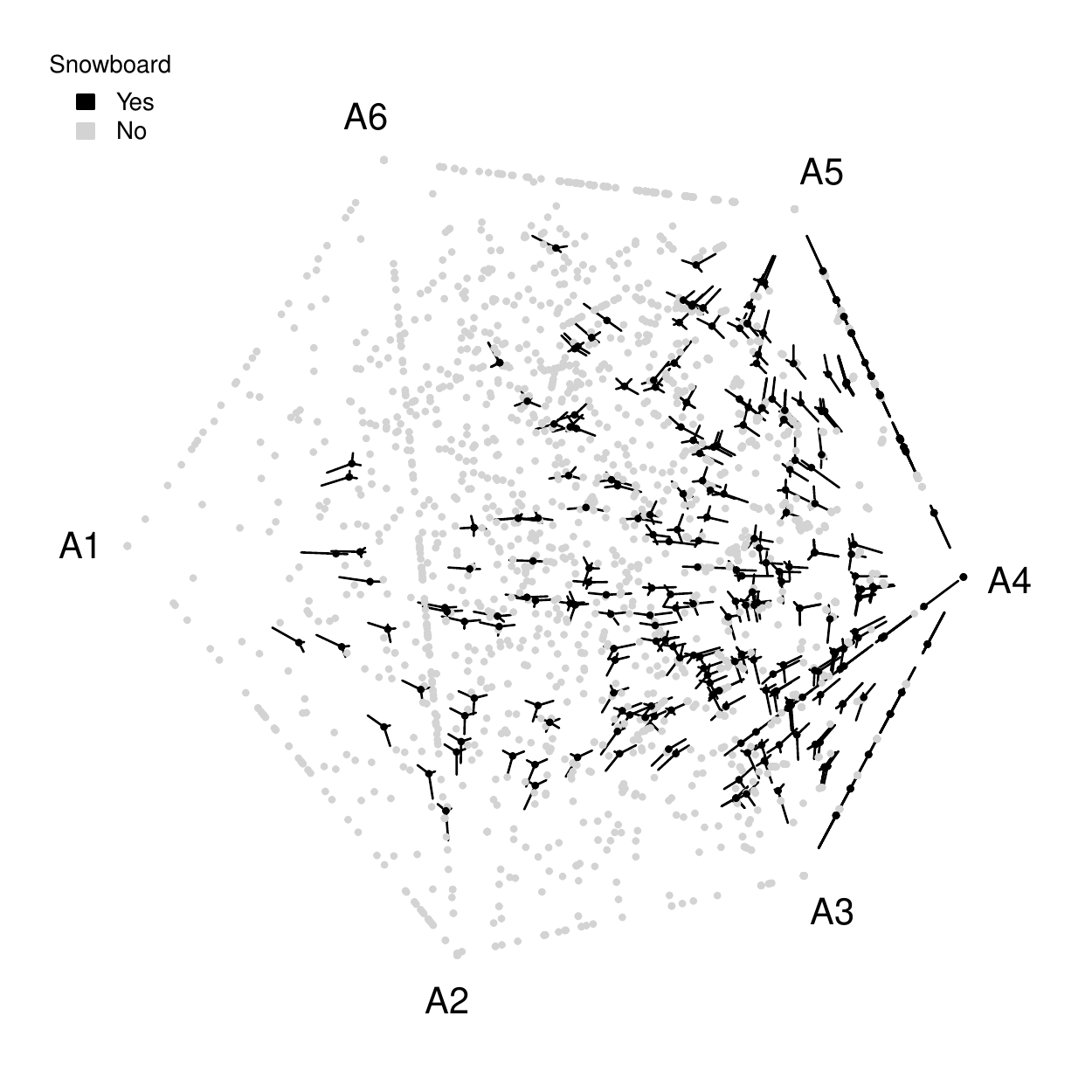}}\\
  \subfigure[]{\includegraphics[scale=0.5]{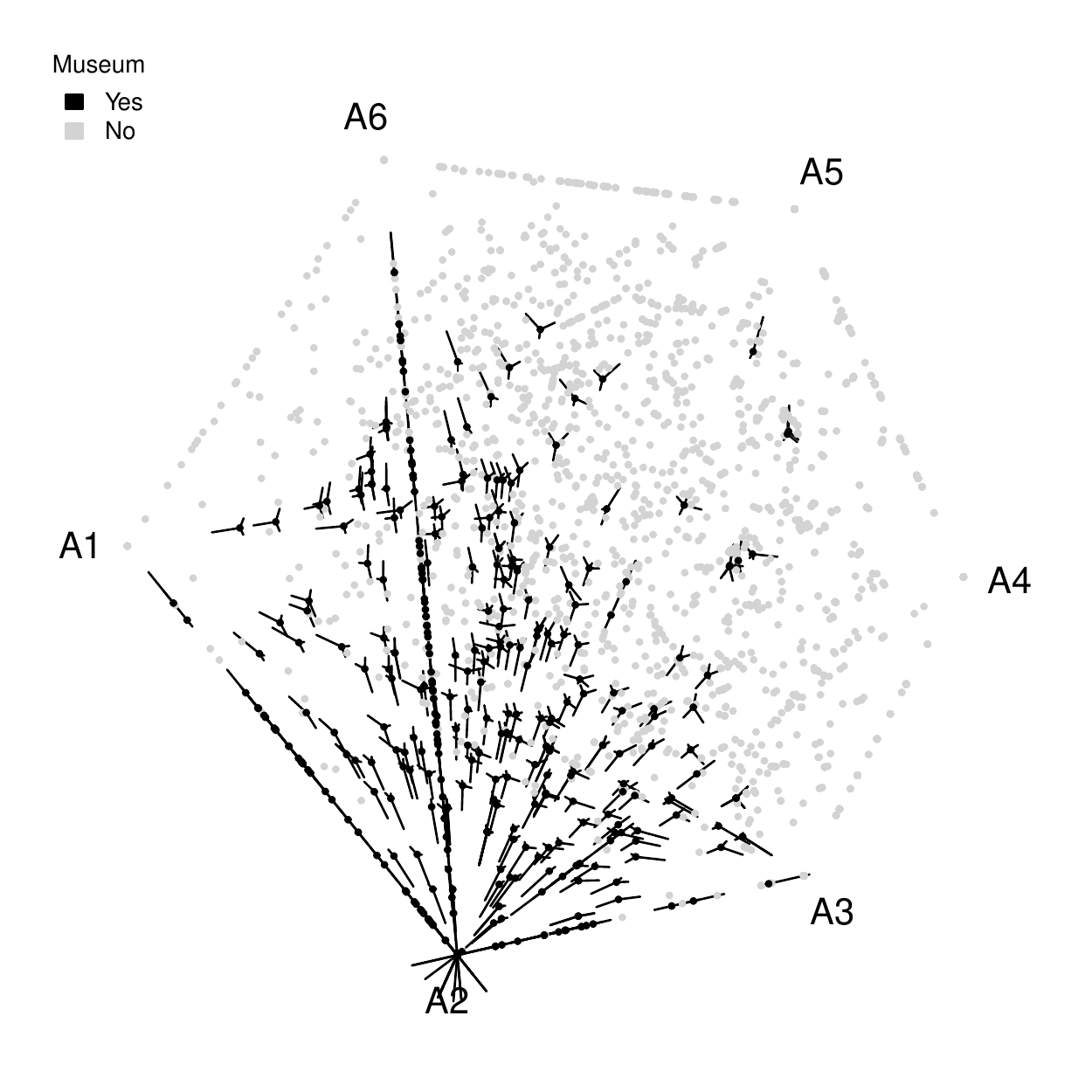}}
  \subfigure[]{\includegraphics[scale=0.5]{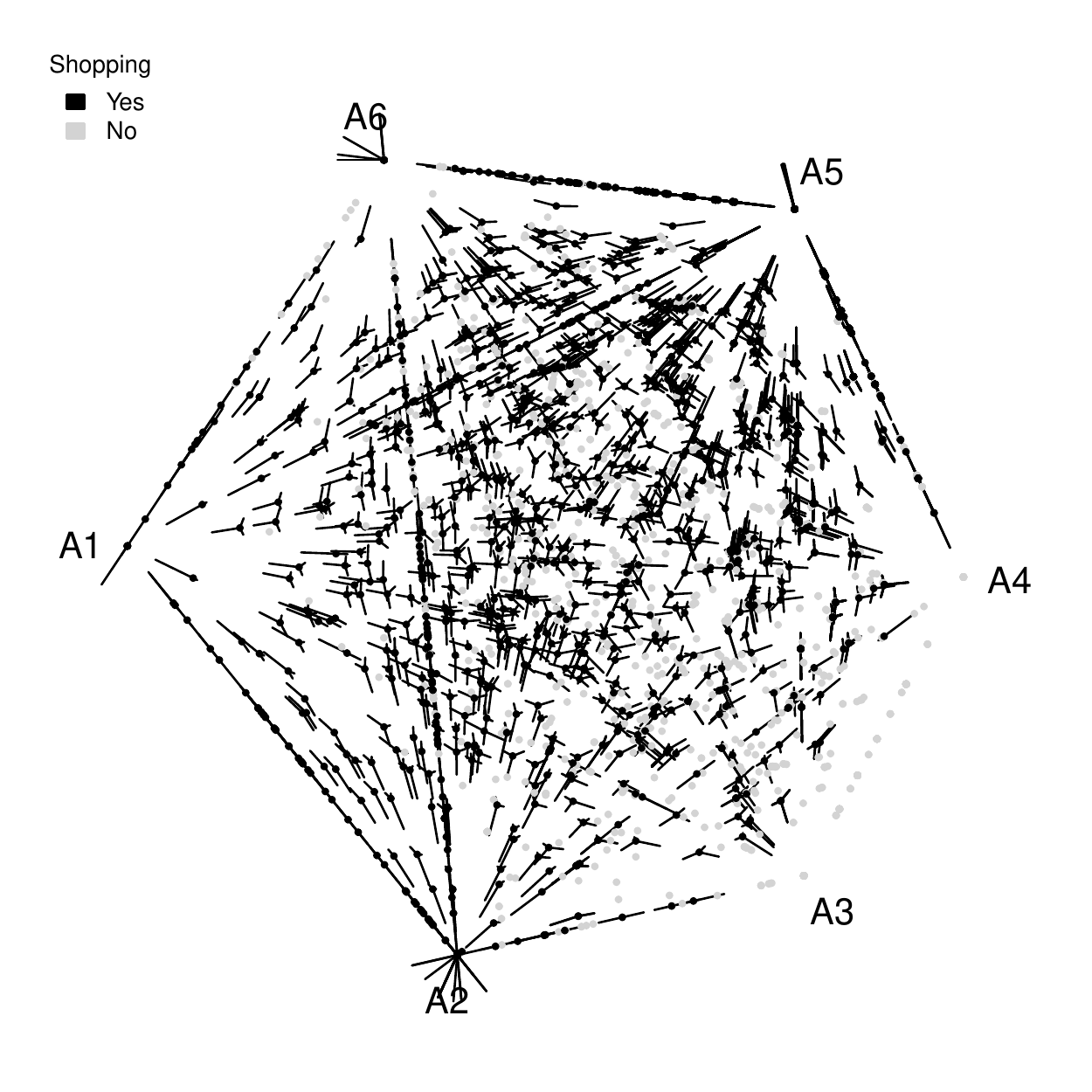}}
  \caption{Simplex visualizations for the the Austrian national guest
    survey example. The archetypes are arranged
    according to their distance in the original space and the
    composition of the winter tourists is indicated by corresponding
    whiskers. Figure~(a) shows the model deviance normalized from 0
    (blue) to 1 (white); figures~(b-d) highlight tourists' answers to
    certain questions (yes/black and no/gray). See
    Section~\ref{subsec:tourist} for detailed interpretations.}
  \label{fig:gsaw97-simplex}
\end{figure*}

To get further insight into the archetypes we explore the simplex visualization.
Figure~\ref{fig:gsaw97-simplex} shows four simplex visualizations with
the archetypes arranged according to their distance in the original
space and the composition of the winter tourists indicated by
corresponding whiskers. Figure~\ref{fig:gsaw97-simplex}a shows the
model deviance normalized from 0 (blue) to 1 (white). We can observe
that the tourists mainly explained by the Minimal~(A3), Modern~(A4),
Traditional~(A5), and/or Wellness~(A6) archetypes are well represented
(darker blue). The Maximal~(A1) archetype seems to be an outlier,
there are only a few observations near to this archetype, and the
deviance is higher for these
observations. Figures~\ref{fig:gsaw97-simplex}b-d highlight
tourists' answers to certain questions (yes/black and no/gray).
Figure~\ref{fig:gsaw97-simplex}b shows whether a tourist does
snowboarding or not. We can see that most of the tourists who do
snowboarding are arranged around and point towards A4, which we
interpreted as the Modern archetype.  In
Figure~\ref{fig:gsaw97-simplex}c we highlight whether a tourist visits
a museum or not. Here, most of the tourists who go there are arranged
around A2, which we interpreted as the Cultural archetype.
Figure~\ref{fig:gsaw97-simplex}d shows an activity which does not
discriminate between archetypes---nearly all tourists do shopping, and
no specific pattern is visible in this visualization.

\newpage

\subsection{Poisson observations: Disasters worldwide from 1900--2008} 
\label{subsec:disaster}

\begin{figure*}[h]
  \centering
\begin{tabular}{cc}
  \multirow{3}{*}[2.5cm]{\includegraphics[scale=0.75]{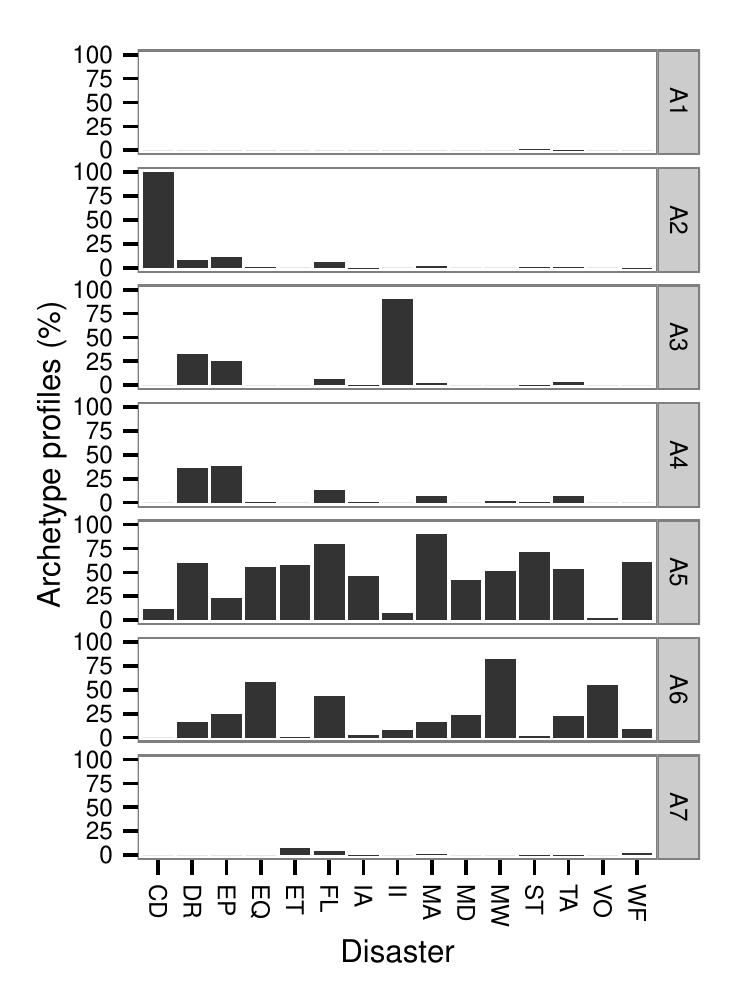}} &
  \includegraphics[  scale=0.5, clip, trim=0.7cm 4.1cm 0.7cm 3.5cm]{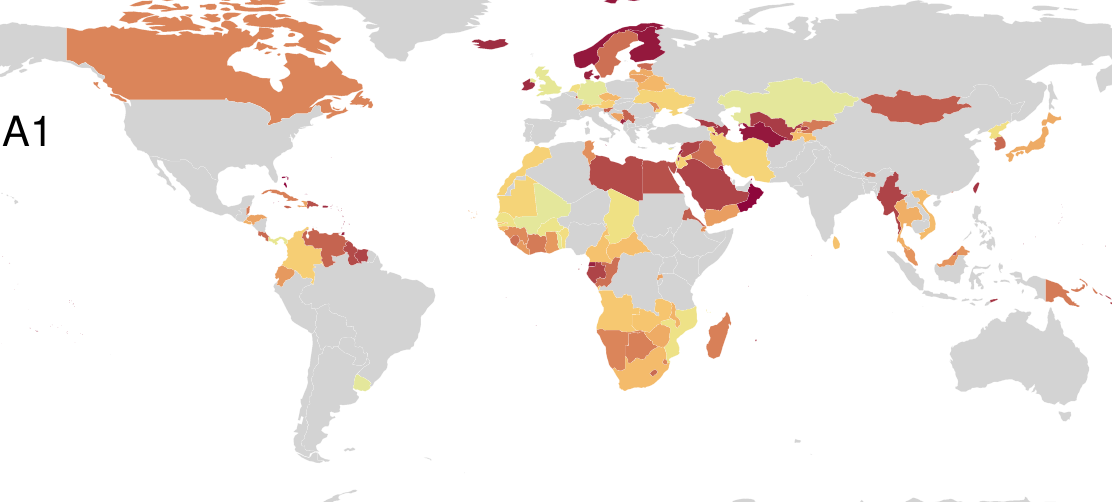} \\
  & \includegraphics[scale=0.5, clip, trim=0.7cm 4.2cm 0.7cm 3.5cm]{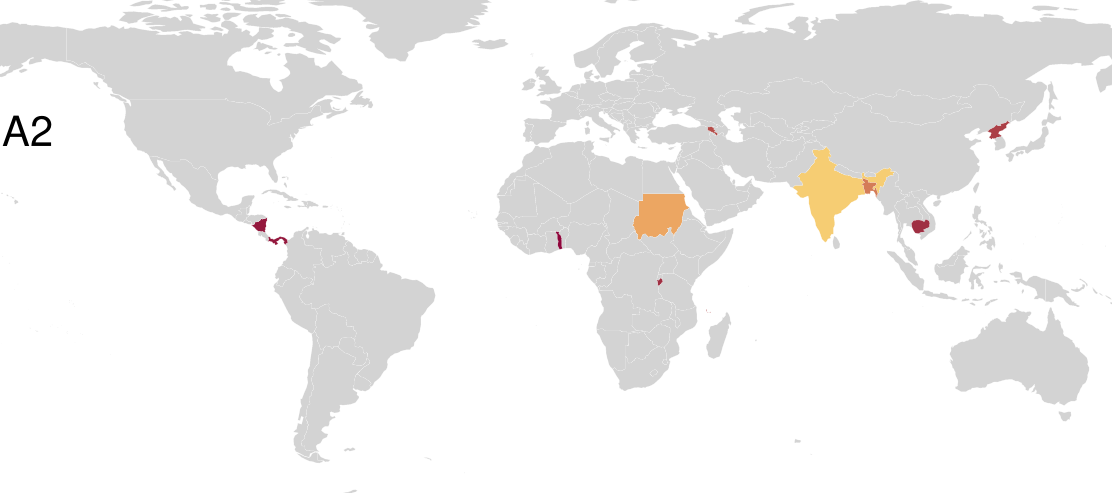} \\
  & \includegraphics[scale=0.5, clip, trim=0.7cm 4.2cm 0.7cm 3.5cm]{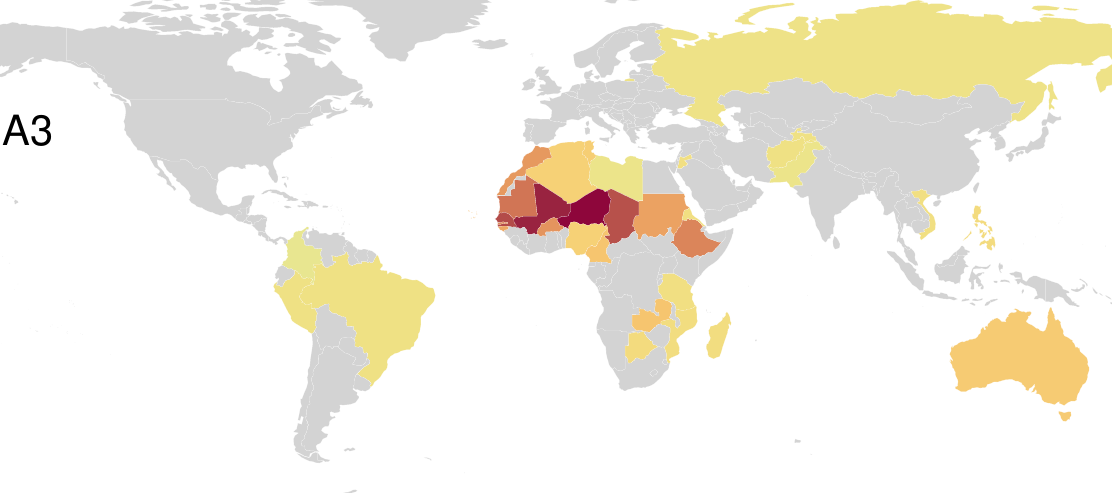} \\
  \includegraphics[  scale=0.5, clip, trim=0.7cm 4.2cm 0.7cm 3.5cm]{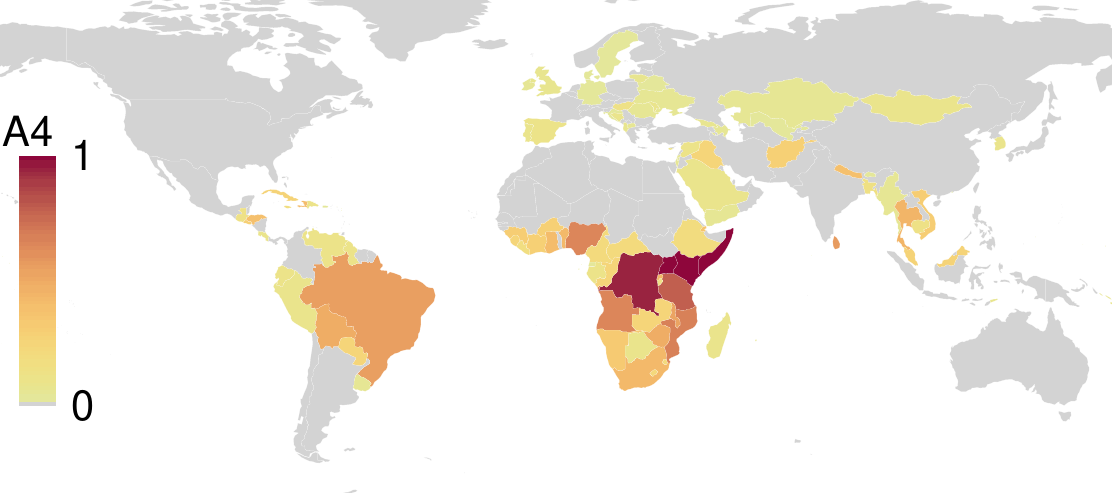} &
  \includegraphics[  scale=0.5, clip, trim=0.7cm 4.2cm 0.7cm 3.5cm]{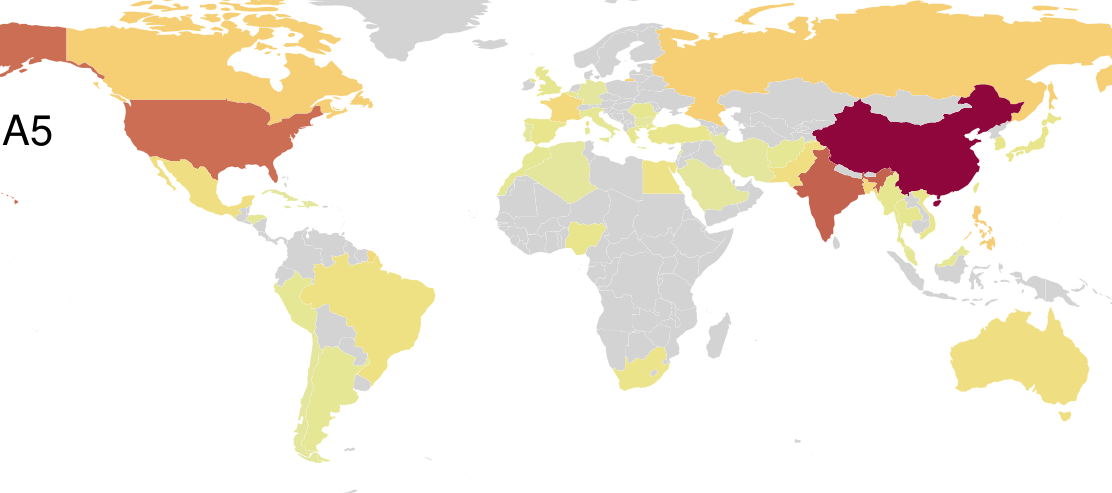} \\
  \includegraphics[  scale=0.5, clip, trim=0.7cm 4.2cm 0.7cm 3.5cm]{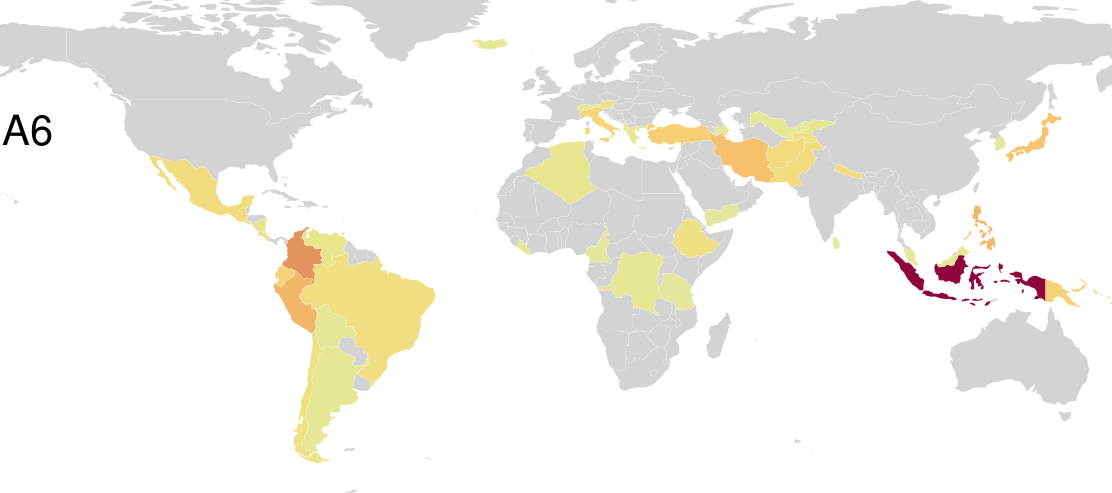} &
  \includegraphics[  scale=0.5, clip, trim=0.7cm 4.2cm 0.7cm 3.5cm]{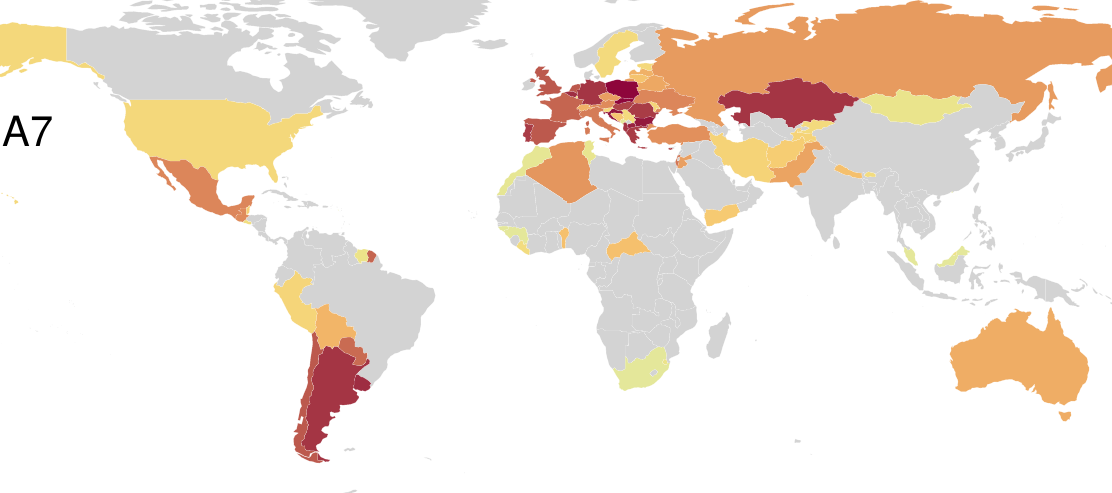} \\
\end{tabular}
\caption{The seven archetypal profiles for the disaster example: (top
  left)~Plot of archetypal profiles ($\%$ of maximum value); (world
  maps) Factors~$\b{H}$ for each archetype. Disasters: complex
  disasters~(CD), drought~(DR), earthquake~(EQ), epidemic~(EP), extreme
  temperature~(ET), flood~(FL), industrial accident~(IA), insect
  infestation~(II), mass movement dry~(MD), mass movement wet~(MW),
  miscellaneous accident~(MA), storm~(ST), transport accident~(TA),
  volcano~(VO), and wildfire~(WF). See Section~\ref{subsec:disaster} for
  details.}
  \label{fig:disaster}
\end{figure*}

In this application, the goal is to identify archetypal countries that
are affected by a particular disaster or a combination of
disasters. This may be helpful in emergency management and to
facilitate devising disaster prevention plans for countries based on
prevention plans designed for the archetypal disaster-affected
countries. We compile a dataset with disaster counts for $227$
countries (historical and present countries) in $15$ categories from
the EM-DAT database~\citep{EM-DAT}. This is a global database on natural and
technological disasters between 1900--present. The criteria to be a
disaster are: ten or more reported casualty; hundred or more people
reported affected; declaration of a state of emergency; or call for
international assistance. The list of disaster categories is provided
in Figure~\ref{fig:disaster}; see the EM-DAT website for specific
details.

\newpage

We present the seven archetypes solution; Figure~\ref{fig:disaster}
shows a summary. There are two minimal profiles A1 and A7 with small
differences in the categories extreme temperature/flood and storm. A1
can be considered as the archetypal profile for safe country where the
corresponding archetypal observations include Malta and the Cayman Islands
(other close observations are the Nordic countries). Archetype A5 is
the maximal archetypal profile with counts in every category, and the
corresponding archetypal observations include China and United
States. This can be expected from the size and population of the
countries; China (third and first), USA (fourth and third). Other
countries with high factor $\b{H}$ for this archetypal profile are
India (seventh and second), and Russia (first and ninth). A3 and A4
are the archetypes that are affected by drought and epidemic; where A3
additionally has a high insect infestation count. A2 is the archetype
that is susceptible to complex disasters (where neither nature nor
human is the definitive cause) only, whereas A6 has high counts in the
categories earthquake, flood, mass movement wet, and volcano.
Here the archetypal countries include Indonesia and Colombia.

\begin{figure}[t]
\centering
  \subfigure[]{\includegraphics[scale=0.5]{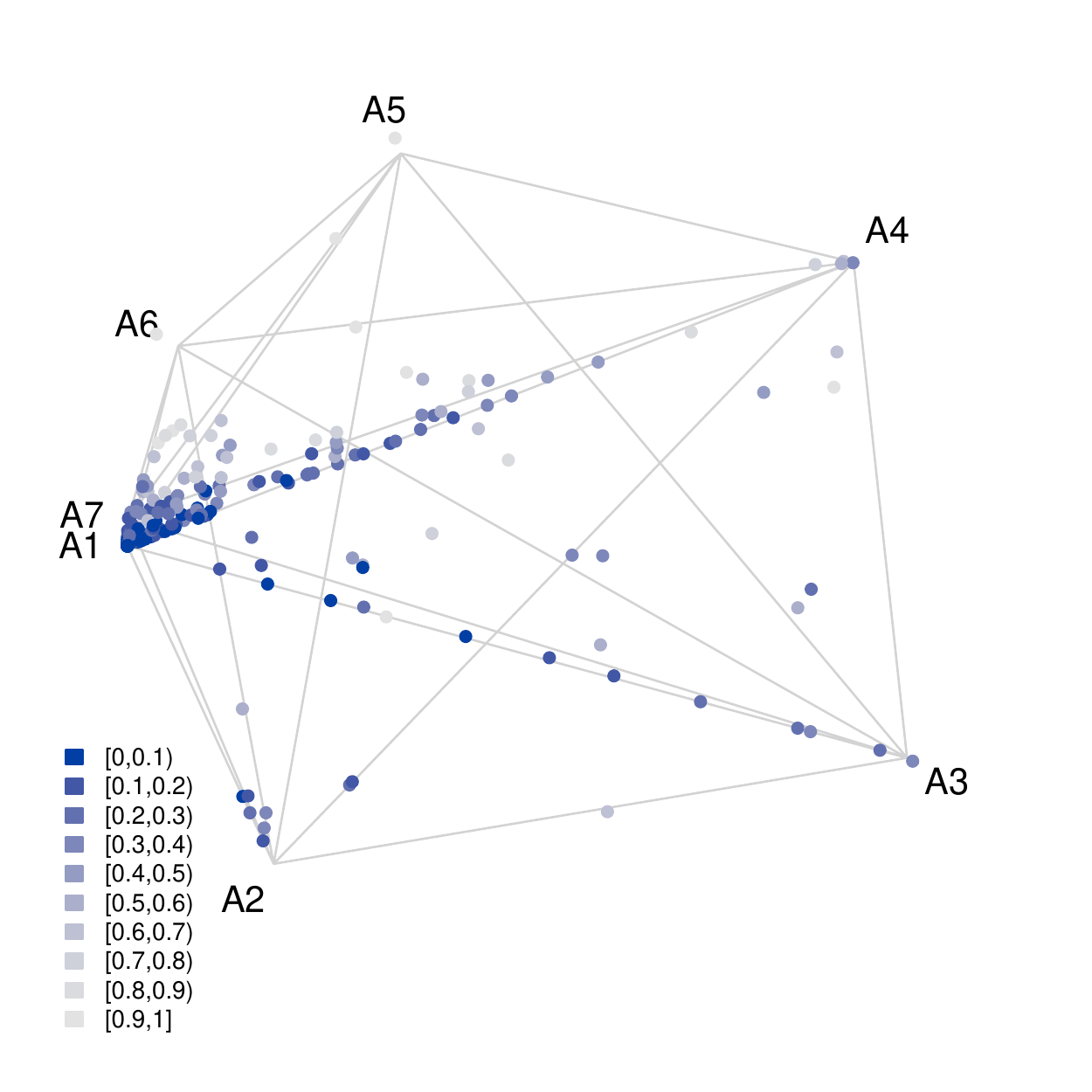}}%
  \subfigure[]{\includegraphics[scale=0.5]{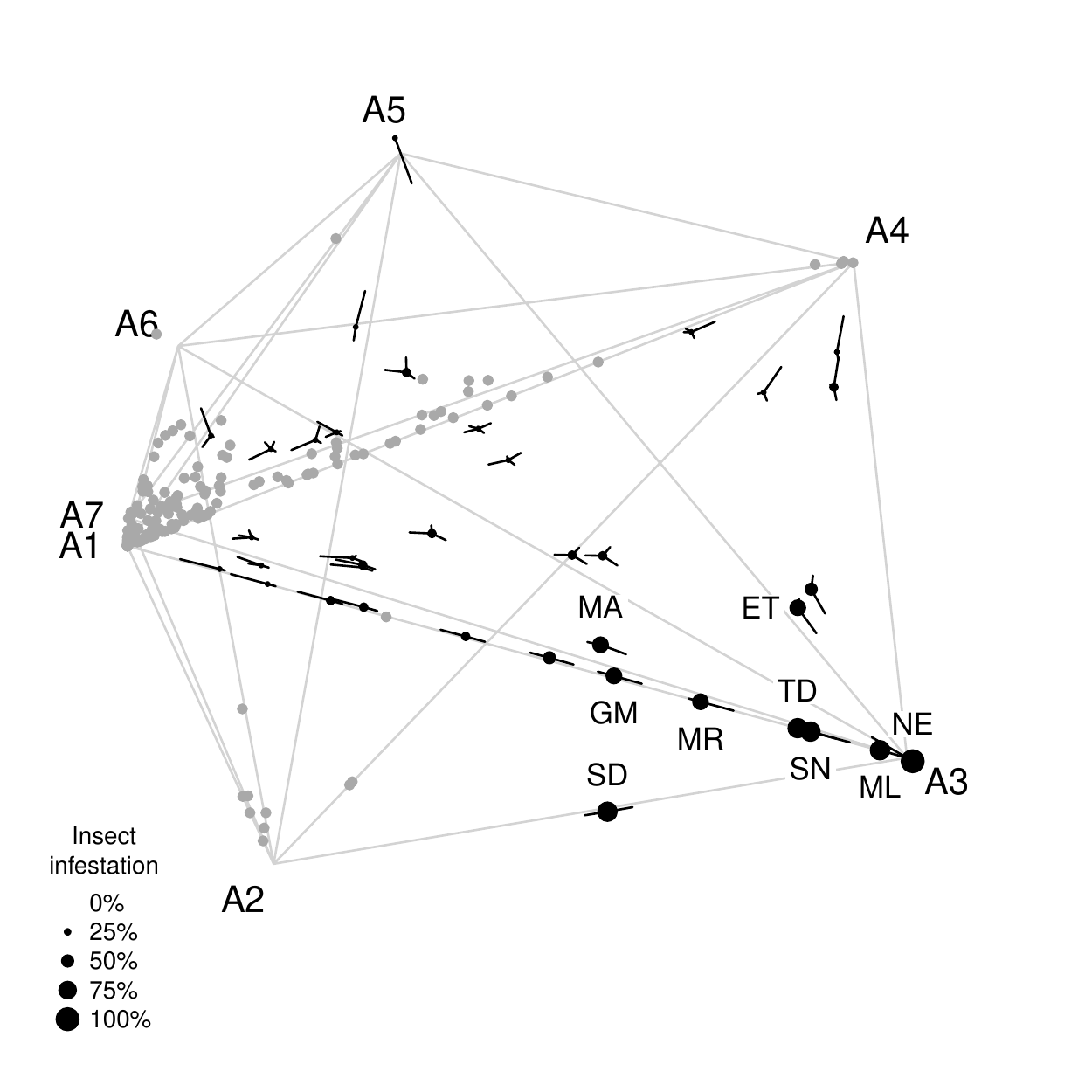}}
  \caption{Simplex visualizations for the disasters example: The
    archetypes are arranged according to their distance in the
    original space. Figure~(a)~ shows the model deviance normalized from 0
    (blue) to 1 (white); Figure~(b) shows the projected countries
    scaled according to the number of Insect infestations (ISO2
    country codes for the top countries). See
    Section~\ref{subsec:disaster} for detailed interpretations.}
  \label{fig:disaster-simplex}
\end{figure}

Figure~\ref{fig:disaster-simplex} shows two simplex visualizations
with the archetypes arranged according to their distance in the
original space. This arrangement shows that A1 is very near to A7,
which is in line with the archetypal profiles plot shown in
Figure~\ref{fig:disaster}. Figure~\ref{fig:disaster-simplex}a shows
the model deviance normalized from 0 (blue) to 1 (white). We can see
that A2, A3, A4, A5 and A6 are basically outliers with only one or
very few observations are around them. Most of the observations with
low deviance are near A1 and A7. Figure~\ref{fig:disaster-simplex}b
highlights the countries' insect infestation (the higher the count the
bigger the point). We can see a clear pattern around A3, with Niger~(NE),
Chad~(TD), Mali~(ML), Senegal~(SN), Sudan~(SD), Ethiopia~(ET),
Gambia~(GM), Mauritania~(MR), and Morocco~(MA) as the top countries
affected by insect infestation.

\section{Discussion}\label{sec:discussion}

Archetypal analysis expresses observations as composition of extreme values, or
archetypes. Archetypes can be thought of as ideal or pure characteristics, and
the goal of archetypal analysis is to find these characteristics, and to
explain the available observations as combination of these characteristics.
The standard formulation of archetypal analysis was suggested by
Cutler and Breiman and is based on finding the approximate convex hull
of the observations. Over the last decade this approach has been
extensively used by researchers.
But, their 
applications have mostly been limited to real valued
observations. In this paper, we have 
proposed a probabilistic formulation of archetypal analysis, which enjoys
several crucial advantages over the geometric approach, including but not
limited to the extension to other observation models: Bernoulli, Poisson and
multinomial.  We have achieved this by approximating the convex hull in the
parameter space under a suitable observation model. Our contribution lies in
formally extending the standard AA framework, suggesting efficient optimization
tools based on majorization-minimization method, and demonstrating the
applicability of such approaches in practical applications. 
We have also suggested improvements of the standard simplex visualization tool
to better show the intricacies in the archetypal analysis solution. 

The probabilistic framework provides further advantages that remains to be
explored in its entirety.  For example, it provides a theoretically sound
approach for choosing the number of archetypes. This can be done by imposing
appropriate prior over $\b{W}$ and $\b{H}$ matrices, such as a symmetric
Dirichlet distribution with coefficient $< 1$.  The prior can be used to
effectively shrink and expand the convex hull to fit the
observations. Since the Dirichlet distribution is a natural prior for
multinomial distribution, this solution can be approximated relatively
easily using variational Bayes' approach, and initial results show
that this is indeed and effective approach for choosing the number of
archetypes. However, this becomes a slightly trickier problem when
applied to other observation models, such as normal and Poisson. We
are currently working on suitable methods to solve the related
optimization problems efficiently.

Another potential extension of the probabilistic framework is to tackle
ordinal or Likert scale variables. Since ordinal variables lack 
additivity, they must be addressed through a probabilistic set-up with
suitable observation model. Given the fact that survey data is often
in Likert scale, archetypal analysis of such observations can have a
large impact on social science and marketing:
describe, e.g., the personality of consumers in terms of the
personality of the most ``extreme'', i.e., archetypal, consumers
(using, e.g., the Likert scaled items defined by the Big Five
Inventory). We believe that these suggested improvements will make
archetypal analysis more robust and accessible to non-scientific
users.

\paragraph{Acknowledgement:}
The calculations presented above were performed using computer
resources within the Aalto University School of Science ``Science-IT''
project.

\appendix

\section{Update rules for multinomial observations}
\label{sec:multi}
For simplicity, let us consider the following problem of finding, $\b{X}=\b{H}\b{W}\b{P}$ where $\b{X}$ is now $n
\times m$ matrix instead of $m \times n$ matrix in the earlier sections. Then this problem can be viewed as given a document choose a topic following $\b{H}$, then given a topic choose a document (subtopic) following $\b{W}$, and finally given a document (subtopic) choose a word following $\b{P}$. Let $\b{R}^{jk}_{il}$ be the indicator variable for selecting topic $j$ and document (subtopic) $k$. Then the log-likelihood of the observations $\b{X}$ is given by
\begin{align*}
\mathbb{LL}\left(\b{X}|\b{H},\b{W},\b{P},\b{R}\right)
&= \sum_{il} \b{X}_{il} \log\left( \prod_{jk} \b{H}_{ij}\b{W}_{jk} \b{P}_{kl} \right)^{z^{jk}_{il}} + C_0 \\
&= \sum_{ijkl} \b{X}_{il}{z^{jk}_{il}}\log\left( \b{H}_{ij}\b{W}_{jk} \b{P}_{kl} \right) + C_0
\end{align*}
At each expectation step we need to evaluate,
\begin{align*}
\E{z^{jk}_{il}|\b{X},\b{H},\b{W},\b{P}} 
= \Prob(z^{jk}_{ik} = 1|\b{X},\b{H},\b{W},\b{P}) 
= \frac{\b{H}_{ij}\b{W}_{jk}\b{P}_{kl}}{(\b{H}\b{W}\b{P})_{il}}
\end{align*}
then the maximization step gives  us the final update equation.

\section{Update rule for Poisson observations}
\label{sec:pois}
We show that the update rule discussed in the article leads to monotonic decrease in the cost function using majorization-minimization. We reformulate the problem as
\begin{align*}
\min_{\b{W},\b{H} \geq 0} &\sum_{ij} \left[ -\b{X}_{ij}\log\sum_{mn } \bs{\Lambda}_{im} \b{W}_{mn} \b{H}_{nj} + \sum_{mn } \bs{\Lambda}_{im} \b{W}_{mn} \b{H}_{nj}\right] \\
&\qquad\qquad+ \lambda\sum_j \left(-\log \sum_n \b{H}_{nj} + \sum_n \b{H}_{nj}  \right) + \lambda\sum_n \left(-\log \sum_m \b{W}_{mn} + \sum_m \b{W}_{mn}  \right)
\end{align*}
where $\lambda > 0$ is a suitably large regularization parameter to enforce the equality constraint in a relaxed fashion. 

Given $\phi_{inj} = \frac{\sum_m\bs{\Lambda}_{im} \b{W}_{mn} \b{H}_{nj}}{\sum_{mn}\bs{\Lambda}_{im} \b{W}_{mn} \b{H}_{nj}}$, and $\psi_{nj} = \frac{\b{H}_{nj}}{\sum_{n}\b{H}_{nj}}$, we can construct the auxiliary function for $\b{H}$ as, 
\begin{align*}
&\sum_{ij} \left[ -\b{X}_{ij}\log\sum_{mn } \bs{\Lambda}_{im} \b{W}_{mn} \tilde{\b{H}}_{nj} + \sum_{mn } \bs{\Lambda}_{im} \b{W}_{mn} \tilde{\b{H}}_{nj} \right] \\
&\qquad+ \lambda\sum_j \left(-\log \sum_n \tilde{\b{H}}_{nj} + \sum_n \tilde{\b{H}}_{nj}  \right)\\
=& \sum_{ij} \left[ -\b{X}_{ij}\log\sum_{n } \frac{\phi_{inj}}{\phi_{inj}} \sum_m\bs{\Lambda}_{im} \b{W}_{mn} \tilde{\b{H}}_{nj} + \sum_{mn } \bs{\Lambda}_{im} \b{W}_{mn} \tilde{\b{H}}_{nj} \right] \\
&\qquad + \lambda\sum_j \left(-\log \sum_n \frac{\psi_{nj}}{\psi_{nj}} \tilde{\b{H}}_{nj} + \sum_n \tilde{\b{H}}_{nj}  \right) \\
\leq& \sum_{ij} \left[ -\b{X}_{ij}\sum_{n } \phi_{inj} \log \frac{\sum_m\bs{\Lambda}_{im} \b{W}_{mn} \tilde{\b{H}}_{nj}}{\phi_{inj}} + \sum_{mn } \bs{\Lambda}_{im} \b{W}_{mn} \tilde{\b{H}}_{nj} \right] \\
&\qquad + \lambda\left(\sum_{nj} -\psi_{nj}\log \frac{\tilde{\b{H}}_{nj}}{\psi_{nj}} + \sum_{nj} \tilde{\b{H}}_{nj} \right) \\
=& \sum_{ij} \left[ -\b{X}_{ij}\sum_{n } \phi_{inj} \log \tilde{\b{H}}_{nj} + \sum_{mn } \bs{\Lambda}_{im} \b{W}_{mn} \tilde{\b{H}}_{nj} \right] \\
&\qquad+ \lambda\left(\sum_{nj} -\psi_{nj}\log\tilde{\b{H}}_{nj} + \sum_{nj} \tilde{\b{H}}_{nj}\right)  + C
\end{align*}
The derivative is then given by
\begin{align*}
\frac{\partial \cdot}{\partial \tilde{\b{H}}_{nj}} =& - \frac{\sum_i \b{X}_{ij}\phi_{inj}}{\tilde{\b{H}}_{nj}} + \sum_{im}\bs{\Lambda}_{im}\b{W}_{mn} - \frac{\lambda\psi_{nj}}{\tilde{\b{H}}_{nj}} + \lambda\\
=&-\frac{\b{H}_{nj}}{\tilde{\b{H}}_{nj}}\left( \sum_i \frac{\b{X}_{ij}\sum_m\bs{\Lambda}_{im} \b{W}_{mn}}{\sum_{mn}\bs{\Lambda}_{im} \b{W}_{mn} \b{H}_{nj}} + \frac{\lambda}{\sum_n \b{H}_{nj}}\right)  + \left( \sum_{im}\bs{\Lambda}_{im}\b{W}_{mn} + \lambda\right)
\end{align*}
Equating the derivative to zero provides the update rule.

Given $\phi_{imnj} = \frac{\bs{\Lambda}_{im} \b{W}_{mn} \b{H}_{nj}}{\sum_{mn}\bs{\Lambda}_{im} \b{W}_{mn} \b{H}_{nj}}$, and $\psi_{mn} = \frac{\b{W}_{mn}}{\sum_{m}\b{W}_{mn}}$, we can construct the auxiliary function 
for $\b{W}$ as,
 
\begin{align*}
&\sum_{ij} \left[ -\b{X}_{ij}\log\sum_{mn } \bs{\Lambda}_{im} \tilde{\b{W}}_{mn} \b{H}_{nj} + \sum_{mn } \bs{\Lambda}_{im} \tilde{\b{W}}_{mn} \b{H}_{nj} \right] \\
&\qquad+ \lambda\sum_n \left(-\log \sum_m \tilde{\b{W}}_{mn} + \sum_m \tilde{\b{W}}_{mn}  \right)\\
=& \sum_{ij} \left[ -\b{X}_{ij}\log\sum_{mn } \frac{\phi_{imnj}}{\phi_{imnj}} \bs{\Lambda}_{im} \tilde{\b{W}}_{mn} \b{H}_{nj} + \sum_{mn } \bs{\Lambda}_{im} \tilde{\b{W}}_{mn} \b{H}_{nj} \right] \\
&\qquad + \lambda\sum_n \left(-\log \sum_m \frac{\psi_{mn}}{\psi_{mn}} \tilde{\b{W}}_{mn} + \sum_m \tilde{\b{W}}_{mn}  \right) \\
\leq& \sum_{ij} \left[ -\b{X}_{ij}\sum_{mn } \phi_{imnj} \log \frac{\bs{\Lambda}_{im} \tilde{\b{W}}_{mn} \b{H}_{nj}}{\phi_{imnj}} + \sum_{mn } \bs{\Lambda}_{im} \tilde{\b{W}}_{mn} \b{H}_{nj} \right] \\
&\qquad + \lambda\left(\sum_{mn} -\psi_{mn}\log \frac{\tilde{\b{W}}_{mn}}{\psi_{mn}} + \sum_{mn} \tilde{\b{W}}_{mn} \right) \\
=& \sum_{ij} \left[ -\b{X}_{ij}\sum_{mn } \phi_{imnj} \log \tilde{\b{W}}_{mn} + \sum_{mn } \bs{\Lambda}_{im} \tilde{\b{W}}_{mn} \b{H}_{nj} \right] \\
&\qquad + \lambda\left(\sum_{mn} -\psi_{mn}\log\tilde{\b{W}}_{mn} + \sum_{mn} \tilde{\b{W}}_{mn}\right)  + C
\end{align*}
The derivative is then given by
\begin{align*}
\frac{\partial \cdot}{\partial \tilde{\b{W}}_{mn}} =& - \frac{\sum_{ij} \b{X}_{ij}\phi_{imnj}}{\tilde{\b{W}}_{mn}} + \sum_{ij}\bs{\Lambda}_{im}\b{H}_{nj} - \frac{\lambda\psi_{mn}}{\tilde{\b{W}}_{mn}} + \lambda\\
=&-\frac{\b{W}_{mn}}{\tilde{\b{W}}_{mn}}\left( \sum_{ij} \frac{\b{X}_{ij}\bs{\Lambda}_{im} \b{H}_{nj}}{\sum_{mn}\bs{\Lambda}_{im} \b{W}_{mn} \b{H}_{nj}} + \frac{\lambda}{\sum_m \b{W}_{mn}}\right)  + \left( \sum_{ij}\bs{\Lambda}_{im}\b{H}_{nj} + \lambda\right)
\end{align*}
Equating the derivative to zero provides the update rule.

\section{Update rule for Bernoulli observations}
\label{sec:bern}

Since the cost function consists of two similar terms, we show how to establish the auxiliary function for one of them.  

For $\b{H}$ we have, $\phi_{inj} = \frac{\sum_{m} \b{P}_{im}\b{W}_{mn}\b{H}_{nj}}{\sum_{mn} \b{P}_{im}\b{W}_{mn}\b{H}_{nj}}$, and $\sum_n \phi_{inj} = 1$, then
\begin{align*}
&\sum_{ij} \left[ - \b{X}_{ij} \log(\b{P}\b{W}\tilde{\b{H}})_{ij} \right] \\
=& \sum_{ij} \left[ - \b{X}_{ij} \log\sum_{mn} \b{P}_{im}\b{W}_{mn}\tilde{\b{H}}_{nj} \right] \\
=& \sum_{ij} \left[ - \b{X}_{ij} \log\sum_{n} \frac{\phi_{inj}}{\phi_{inj}} \sum_{m} \b{P}_{im}\b{W}_{mn}\tilde{\b{H}}_{nj} \right] \\
\leq & \sum_{ij} \left[ - \b{X}_{ij} \sum_{n} \phi_{inj}  \log \frac{\sum_{m} \b{P}_{im}\b{W}_{mn}\tilde{\b{H}}_{nj}}{\phi_{inj}} \right] \\
=& \sum_{ij} \left[ - \b{X}_{ij} \sum_{n} \phi_{inj}  \log \frac{\tilde{\b{H}}_{nj}}{\b{H}_{nj}} - \b{X}_{ij} \log\sum_{mn} \b{P}_{im}\b{W}_{mn}\b{H}_{nj}\right]\\
=& \sum_{ij} \left[ - \b{X}_{ij} \sum_{n} \phi_{inj}  \log \left( \frac{\tilde{\b{G}}_{nj}}{\sum_p\tilde{\b{G}}_{pj}} \frac{\sum_p\b{G}_{pj}}{\b{G}_{nj}}\right) - \b{X}_{ij} \log\sum_{mn} \b{P}_{im}\b{W}_{mn}\b{H}_{nj}\right]\\
=& \sum_{ij} \left[ - \b{X}_{ij} \sum_{n} \phi_{inj}  \log \frac{\tilde{\b{G}}_{nj}}{\b{G}_{nj}} + \b{X}_{ij} \sum_{n} \phi_{inj}  \log \frac{\sum_p\tilde{\b{G}}_{pj}}{\sum_p\b{G}_{pj}} - \b{X}_{ij} \log\sum_{mn} \b{P}_{im}\b{W}_{mn}\b{H}_{nj}\right]\\
\leq& \sum_{ij} \left[ - \b{X}_{ij} \sum_{n} \phi_{inj}  \log \frac{\tilde{\b{G}}_{nj}}{\b{G}_{nj}} + \b{X}_{ij} \left( \frac{\sum_p\tilde{\b{G}}_{pj}}{\sum_p\b{G}_{pj}} - 1\right)- \b{X}_{ij} \log\sum_{mn} \b{P}_{im}\b{W}_{mn}\b{H}_{nj}\right]
\end{align*}
Taking derivative we get,
\begin{align*}
&\sum_{i} \left[ - \b{X}_{ij} \phi_{inj}  \frac{1}{\tilde{\b{G}}_{nj}} + \b{X}_{ij} \frac{1}{\sum_p\b{G}_{pj}} \right] \\
\Rightarrow& \tilde{\b{G}}_{nj} = \frac{\sum_p \b{G}_{pj}}{\sum_i \b{X}_{ij}}\left( \sum_i \frac{\b{X}_{ij}\sum_{m} \b{P}_{im}\b{W}_{mn}\b{H}_{nj}}{\sum_{mn} \b{P}_{im}\b{W}_{mn}\b{H}_{nj}}\right) = \frac{\b{G}_{nj}}{\sum_i \b{X}_{ij}}\left( \sum_i \frac{\b{X}_{ij}\sum_{m} \b{P}_{im}\b{W}_{mn}}{\sum_{mn} \b{P}_{im}\b{W}_{mn}\b{H}_{nj}}\right)
\end{align*}

For $\b{W}$ we have, $\phi_{imnj} = \frac{\b{P}_{im}\b{W}_{mn}\b{H}_{nj}}{\sum_{mn} \b{P}_{im}\b{W}_{mn}\b{H}_{nj}}$, and $\sum_{mn} \phi_{imnj} = 1$, then
\begin{align*}
&\sum_{ij} \left[ - \b{X}_{ij} \log(\b{P}\tilde{\b{W}}\b{H})_{ij} \right] \\
=& \sum_{ij} \left[ - \b{X}_{ij} \log\sum_{mn} \b{P}_{im}\tilde{\b{W}}_{mn}\b{H}_{nj} \right] \\
=& \sum_{ij} \left[ - \b{X}_{ij} \log\sum_{mn} \frac{\phi_{imnj}}{\phi_{imnj}} \b{P}_{im}\tilde{\b{W}}_{mn}\b{H}_{nj} \right] \\
\leq & \sum_{ij} \left[ - \b{X}_{ij} \sum_{mn} \phi_{imnj}  \log \frac{\b{P}_{im}\tilde{\b{W}}_{mn}\b{H}_{nj}}{\phi_{imnj}} \right] \\
=& \sum_{ij} \left[ - \b{X}_{ij} \sum_{mn} \phi_{imnj}  \log \frac{\tilde{\b{W}}_{mn}}{\b{W}_{mn}} - \b{X}_{ij} \log\sum_{mn} \b{P}_{im}\b{W}_{mn}\b{H}_{nj}\right]\\
=& \sum_{ij} \left[ - \b{X}_{ij} \sum_{mn} \phi_{imnj}  \log \left( \frac{\tilde{\b{V}}_{mn}}{\sum_p\tilde{\b{V}}_{pn}} \frac{\sum_p\b{V}_{pn}}{\b{V}_{mn}}\right) - \b{X}_{ij} \log\sum_{mn} \b{P}_{im}\b{W}_{mn}\b{H}_{nj}\right]\\
=& \sum_{ij} \left[ - \b{X}_{ij} \sum_{mn} \phi_{imnj}  \log \frac{\tilde{\b{V}}_{mn}}{\b{V}_{mn}} + \b{X}_{ij} \sum_{mn} \phi_{imnj}  \log \frac{\sum_p\tilde{\b{V}}_{pn}}{\sum_p\b{V}_{pn}} - \b{X}_{ij} \log\sum_{mn} \b{P}_{im}\b{W}_{mn}\b{H}_{nj}\right]\\
\leq& \sum_{ij} \left[ - \b{X}_{ij} \sum_{mn} \phi_{imnj}  \log \frac{\tilde{\b{V}}_{mn}}{\b{V}_{mn}} + \b{X}_{ij} \sum_{mn} \phi_{imnj} \left( \frac{\sum_p\tilde{\b{V}}_{pn}}{\sum_p\b{V}_{pn}} - 1\right)- \b{X}_{ij} \log\sum_{mn} \b{P}_{im}\b{W}_{mn}\b{H}_{nj}\right]
\end{align*}
Taking derivative we get,
\begin{align*}
&\sum_{ij} \left[ - \b{X}_{ij} \phi_{imnj}  \frac{1}{\tilde{\b{V}}_{mn}} + \b{X}_{ij} \sum_m \phi_{imnj}\frac{1}{\sum_p\b{V}_{pn}} \right] \\
\Rightarrow& \tilde{\b{V}}_{mn} = \frac{\sum_p \b{V}_{pn}}{\sum_{ij} \b{X}_{ij}\sum_m \phi_{imnj}}\left( \sum_{ij} \frac{\b{X}_{ij}\b{P}_{im}\b{W}_{mn}\b{H}_{nj}}{\sum_{mn} \b{P}_{im}\b{W}_{mn}\b{H}_{nj}}\right) \\
& \qquad =\frac{\b{V}_{mn}}{\sum_{ij} \frac{\b{X}_{ij}\sum_m \b{P}_{im}\b{W}_{mn}\b{H}_{nj}}{\sum_{mn} \b{P}_{im}\b{W}_{mn}\b{H}_{nj}} }\left( \sum_{ij} \frac{\b{X}_{ij}\b{P}_{im}\b{H}_{nj}}{\sum_{mn} \b{P}_{im}\b{W}_{mn}\b{H}_{nj}}\right)
\end{align*}
The update rule can be derived from these equations after including the other term with $\b{Q}$.


\begin{thebibliography}{38}
	\providecommand{\natexlab}[1]{#1}
	\providecommand{\url}[1]{{#1}}
	\providecommand{\urlprefix}{URL }
	\expandafter\ifx\csname urlstyle\endcsname\relax
	  \providecommand{\doi}[1]{DOI~\discretionary{}{}{}#1}\else
	  \providecommand{\doi}{DOI~\discretionary{}{}{}\begingroup
  \urlstyle{rm}\Url}\fi
\providecommand{\eprint}[2][]{\url{#2}}

\bibitem[{Bache and Lichman(2013)}]{Bache+Lichman:2013}
Bache K, Lichman M (2013) {UCI} machine learning repository.
  \urlprefix\url{http://archive.ics.uci.edu/ml}

\bibitem[{Bauckhage and Thurau(2009)}]{Bauckhage+Thurau@2009}
Bauckhage C, Thurau C (2009) Making archetypal analysis practical. In: Pattern
  Recognition, Lecture Notes in Computer Science, vol 5748, Springer Berlin
  Heidelberg, pp 272--281, \doi{10.1007/978-3-642-03798-6_28}

\bibitem[{Bhattacharya and Dunson(2012)}]{bhattacharya_simplex_2012}
Bhattacharya A, Dunson DB (2012) Simplex factor models for multivariate
  unordered categorical data. Journal of the American Statistical Association
  107(497):362--377, \doi{10.1080/01621459.2011.646934}

\bibitem[{Blei et~al(2003)Blei, Ng, and Jordan}]{Blei:2003:LDA}
Blei DM, Ng AY, Jordan MI (2003) Latent dirichlet allocation. Journal of
  Machine Learning Research 3:993--1022

\bibitem[{Chan et~al(2003)Chan, Mitchell, and Cram}]{chan_archetypal_2003}
Chan BHP, Mitchell DA, Cram LE (2003) Archetypal analysis of galaxy spectra.
  Monthly Notices of the Royal Astronomical Society 338(3):790--795,
  \doi{10.1046/j.1365-8711.2003.06099.x}

\bibitem[{Cutler and Breiman(1994)}]{cutler_adele_archetypal_1994}
Cutler A, Breiman L (1994) Archetypal analysis. Technometrics 36(4):338--347

\bibitem[{Cutler and Stone(1997)}]{cutler_moving_1997}
Cutler A, Stone E (1997) Moving archetypes. Physica D: Nonlinear Phenomena
  107(1):1--16, \doi{10.1016/S0167-2789(97)84209-1},
  \urlprefix\url{http://www.sciencedirect.com/science/article/pii/S0167278997842091}

\bibitem[{Davis and Love(2010)}]{Davis+Love@2010}
Davis T, Love BC (2010) Memory for category information is idealized through
  contrast with competing options. Psychological Science 21(2):234--242,
  \doi{10.1177/0956797609357712}

\bibitem[{Ding et~al(2006)Ding, Li, Peng, and Park}]{ding_orthogonal_2006}
Ding C, Li T, Peng W, Park H (2006) Orthogonal nonnegative matrix
  tri-factorizations for clustering. In: Proceedings of the 12th {ACM} {SIGKDD}
  International Conference on Knowledge Discovery and Data Mining, pp 126--135,
  \doi{10.1145/1150402.1150420}

\bibitem[{Ding et~al(2010)Ding, Li, and Jordan}]{Ding:2010}
Ding CHQ, Li T, Jordan MI (2010) Convex and semi-nonnegative matrix
  factorizations. IEEE Transactions on Pattern Analysis and Machine
  Intelligence 32(1):45--55

\bibitem[{Dolnicar et~al(2011)Dolnicar, Gr{\" u}n, and
  Leisch}]{Dolnicar+Grun+Leisch@2011}
Dolnicar S, Gr{\" u}n B, Leisch F (2011) Quick, simple and reliable: Forced
  binary survey questions. International Journal of Market Research
  53(2):231--252, \doi{10.2501/IJMR-53-2-231-252}

\bibitem[{{EM-DAT}(2013)}]{EM-DAT}
{EM-DAT} (2013) The OFDA/CRED International Disaster Database. Universite
  catholique de Louvain, Brussels, Belgium; \texttt{http://www.emdat.net}

\bibitem[{Eugster(2012)}]{Eugster@2012}
Eugster MJA (2012) Performance profiles based on archetypal athletes.
  International Journal of Performance Analysis in Sport 12(1):166--187

\bibitem[{Eugster and Leisch(2011)}]{Eugster+Leisch@2011}
Eugster MJA, Leisch F (2011) Weighted and robust archetypal analysis.
  Computational Statistics and Data Analysis 55(3):1215--1225,
  \doi{10.1016/j.csda.2010.10.017}

\bibitem[{Eugster and Leisch(2013)}]{archetypes}
Eugster MJA, Leisch F (2013) archetypes: {A}rchetypal Analysis.
  \urlprefix\url{http://CRAN.R-project.org/package=archetypes}, {R} package
  version 2.1-2

\bibitem[{F{\'e}votte and Idier(2011)}]{DBLP:journals/neco/FevotteI11}
F{\'e}votte C, Idier J (2011) Algorithms for nonnegative matrix factorization
  with the beta-divergence. Neural Computation 23(9):2421--2456

\bibitem[{Friendly(2000)}]{Friendly:2000}
Friendly M (2000) Visualizing Categorical Data. SAS Institute

\bibitem[{Hahsler and Hornik(2007)}]{Hahsler+Hornik:2007}
Hahsler M, Hornik K (2007) {TSP} -- {I}nfrastructure for the traveling
  salesperson problem. Journal of Statistical Software 23(2):1--21,
  \urlprefix\url{http://www.jstatsoft.org/v23/i02/}

\bibitem[{Hofmann(2013)}]{hofmann_probabilistic_2013}
Hofmann T (2013) Probabilistic latent semantic analysis. {arXiv:13016705}
  \urlprefix\url{http://arxiv.org/abs/1301.6705}

\bibitem[{Lee and Seung(1999)}]{lee_learning_1999}
Lee DD, Seung HS (1999) Learning the parts of objects by non-negative matrix
  factorization. Nature 401(6755):788--791, \doi{10.1038/44565}

\bibitem[{Lee and Seung(2000)}]{lee_algorithms_2000}
Lee DD, Seung HS (2000) Algorithms for non-negative matrix factorization. In:
  Advances in Neural Information Processing Systems 13, pp 556--562

\bibitem[{Li et~al(2003)Li, Louviere, Carson, and Wang}]{li_archetypal_2003}
Li S, Louviere J, Carson R, Wang P (2003) Archetypal analysis: {A} new way to
  segment markets based on extreme individuals. In: A Celebration of Ehrenberg
  and Bass: Marketing Knowledge, Discoveries and Contribution. Proceedings of
  the ANZMAC 2003 Conference,
  \urlprefix\url{http://epress.lib.uts.edu.au/research/handle/10453/2183}

\bibitem[{Marinetti et~al(2007)Marinetti, Finesso, and
  Marsilio}]{marinetti_archetypes_2007}
Marinetti S, Finesso L, Marsilio E (2007) Archetypes and principal components
  of an {IR} image sequence. Infrared Physics \& Technology 49(3):272--276,
  \doi{10.1016/j.infrared.2006.06.017},
  \urlprefix\url{http://www.sciencedirect.com/science/article/pii/S1350449506000910}

\bibitem[{Mohamed et~al(2009)Mohamed, Heller, and
  Ghahramani}]{DBLP:conf/nips/MohamedHG08}
Mohamed S, Heller KA, Ghahramani Z (2009) Bayesian exponential family {PCA}.
  In: Advances in Neural Information Processing Systems 21, pp 1089--1096

\bibitem[{M{\o}rup and Hansen(2012)}]{morup_archetypal_2012}
M{\o}rup M, Hansen LK (2012) Archetypal analysis for machine learning and data
  mining. Neurocomputing 80:54--63, \doi{10.1016/j.neucom.2011.06.033}

\bibitem[{do~Nascimento and Dias(2005)}]{nascimento_vertex_2005}
do~Nascimento JMP, Dias JMB (2005) Vertex component analysis: {A} fast
  algorithm to unmix hyperspectral data. {IEEE} Transactions on Geoscience and
  Remote Sensing 43(4):898--910, \doi{10.1109/TGRS.2005.844293}

\bibitem[{Porzio et~al(2008)Porzio, Ragozini, and Vistocco}]{porzio_use_2008}
Porzio GC, Ragozini G, Vistocco D (2008) On the use of archetypes as
  benchmarks. Applied Stochastic Models in Business and Industry
  24(5):419--437, \doi{10.1002/asmb.727},
  \urlprefix\url{http://onlinelibrary.wiley.com/doi/10.1002/asmb.727/abstract}

\bibitem[{Seiler and Wohlrabe(2013)}]{Seiler+Wohlrabe@2013}
Seiler C, Wohlrabe K (2013) Archetypal scientists. Journal of Informetrics
  7(2):345--356, \doi{10.1016/j.joi.2012.11.013}

\bibitem[{Sifa and Bauckhage(2013)}]{sifa_archetypical_2013}
Sifa R, Bauckhage C (2013) Archetypical motion: {S}upervised game behavior
  learning with archetypal analysis. In: 2013 {IEEE} Conference on
  Computational Intelligence in Games ({CIG)}, pp 1--8,
  \doi{10.1109/CIG.2013.6633609}

\bibitem[{Steinley(2006)}]{Steinley@2006}
Steinley D (2006) K-means clustering: {A} half-century synthesis. British
  Journal of Mathematical and Statistical Psychology 59(1):1--34,
  \doi{10.1348/000711005X48266}

\bibitem[{Stone and Cutler(1996)}]{stone_archetypal_1996}
Stone E, Cutler A (1996) Archetypal analysis of spatio-temporal dynamics.
  Physica D: Nonlinear Phenomena 90(3):209--224,
  \doi{10.1016/0167-2789(95)00244-8}

\bibitem[{Th{\o}gersen et~al(2013)Th{\o}gersen, M{\o}rup, Damki{\ae}r, Molin,
  and Jelsbak}]{thogersen_archetypal_2013}
Th{\o}gersen JC, M{\o}rup M, Damki{\ae}r S, Molin S, Jelsbak L (2013)
  Archetypal analysis of diverse pseudomonas aeruginosa transcriptomes reveals
  adaptation in cystic fibrosis airways. {BMC} Bioinformatics 14(1):279,
  \doi{10.1186/1471-2105-14-279},
  \urlprefix\url{http://www.biomedcentral.com/1471-2105/14/279/abstract}

\bibitem[{Thurau et~al(2009)Thurau, Kersting, and
  Bauckhage}]{thurau_convex_2009}
Thurau C, Kersting K, Bauckhage C (2009) Convex non-negative matrix
  factorization in the wild. In: Ninth {IEEE} International Conference on Data
  Mining, 2009. {ICDM} '09, pp 523--532, \doi{10.1109/ICDM.2009.55}

\bibitem[{Thurau et~al(2010)Thurau, Kersting, and Bauckhage}]{thurau_yes_2010}
Thurau C, Kersting K, Bauckhage C (2010) Yes we can: {S}implex volume
  maximization for descriptive web-scale matrix factorization. In: Proceedings
  of the 19th {ACM} International Conference on Information and Knowledge
  Management, {ACM}, New York, {NY}, {USA}, {CIKM} '10, pp 1785--1788,
  \doi{10.1145/1871437.1871729},
  \urlprefix\url{http://doi.acm.org/10.1145/1871437.1871729}

\bibitem[{Tibshirani and Walther(2005)}]{Tibshirani+Walther@2005}
Tibshirani R, Walther G (2005) Cluster validation by prediction strength.
  Journal of Computational and Graphical Statistics 14:511--528

\bibitem[{Woodbury and Clive(1974)}]{woodbury_clinical_1974}
Woodbury MA, Clive J (1974) Clinical pure types as a fuzzy partition. Journal
  of Cybernetics 4(3):111--121, \doi{10.1080/01969727408621685},
  \urlprefix\url{http://www.tandfonline.com/doi/abs/10.1080/01969727408621685}

\bibitem[{Xiong et~al(2013)Xiong, Liu, Zhao, and Tang}]{xiong_face_2013}
Xiong Y, Liu W, Zhao D, Tang X (2013) Face recognition via archetype hull
  ranking. In: 2013 {IEEE} International Conference on Computer Vision
  ({ICCV)}, pp 585--592, \doi{10.1109/ICCV.2013.78}

\bibitem[{Yang and Oja(2012)}]{yang_clustering_2012}
Yang Z, Oja E (2012) Clustering by low-rank doubly stochastic matrix
  decomposition. {arXiv:12064676}
  \urlprefix\url{http://arxiv.org/abs/1206.4676}

\end{thebibliography}

\end{document}